\documentclass[12pt]{article}%
\usepackage{amsmath}
\usepackage{amsfonts}
\usepackage{amssymb}
\usepackage{graphicx}
\usepackage{hyperref}%
\providecommand{\U}[1]{\protect \rule{.1in}{.1in}}
\providecommand{\U}[1]{\protect \rule{.1in}{.1in}}

\begin{document}

\title{Survival Concept-Based Learning Models}
\author{Stanislav R. Kirpichenko, Lev V. Utkin\\Andrei V. Konstantinov, Natalya M. Verbova \\{\small {Higher School of Artificial Intelligence Technologies} }\\{\small {Peter the Great St.Petersburg Polytechnic University}}\\{\small {St.Petersburg, Russia} } \\{\small e-mail: kirpich\_sr@spbstu.ru, utkin\_lv@spbstu.ru}\\{\small konstantinov\_av@spbstu.ru, verbova\_nm@spbstu.ru}}
\date{}
\maketitle

\begin{abstract}
Concept-based learning enhances prediction accuracy and interpretability by
leveraging high-level, human-understandable concepts. However, existing CBL
frameworks do not address survival analysis tasks, which involve predicting
event times in the presence of censored data -- a common scenario in fields
like medicine and reliability analysis. To bridge this gap, we propose two
novel models: SurvCBM (Survival Concept-based Bottleneck Model) and SurvRCM
(Survival Regularized Concept-based Model), which integrate concept-based
learning with survival analysis to handle censored event time data. The models
employ the Cox proportional hazards model and the Beran estimator. SurvCBM is
based on the architecture of the well-known concept bottleneck model, offering
interpretable predictions through concept-based explanations. SurvRCM uses
concepts as regularization to enhance accuracy. Both models are trained
end-to-end and provide interpretable predictions in terms of concepts. Two
interpretability approaches are proposed: one leveraging the linear
relationship in the Cox model and another using an instance-based explanation
framework with the Beran estimator. Numerical experiments demonstrate that
SurvCBM outperforms SurvRCM and traditional survival models, underscoring the
importance and advantages of incorporating concept information. The code for
the proposed algorithms is publicly available.

\textit{Keywords}: concept-based learning, survival analysis, neural networks,
censored data, interpretation

\end{abstract}

\section{Introduction}

Concept-based learning (CBL)
\cite{Gupta-Narayanan-24,kim2018interpretability,lage2020learning,wang2023learning,xu2023statistically,yeh2020completeness}
is an approach aimed at improving classification or regression accuracy. More
importantly, it addresses the challenge of interpreting predictions generated
by deep learning models
\cite{Guidotti-2019,Liang-etal-2021,Zhang-Tino-etal-2020}. Concepts represent
semantic descriptions of input images and can be viewed as high-level,
human-understandable attributes or abstractions
\cite{Gupta-Narayanan-24,poeta2023concept}. In most cases, concept labels are
annotated by humans or domain experts \cite{hu2024editable}. Unlike
conventional black-box models, which establish a direct relationship between
input data and predictions, CBL models first predict concept values from the
input data and then use these predicted concepts to determine target class
labels \cite{koh2020concept}.

One of the key architectures within the CBL framework is the concept
bottleneck model (CBM), introduced by Koh et al. \cite{koh2020concept}. This
model consists of two predictors: a concept predictor and a class predictor.
The concept predictor explicitly generates concept labels from images, while
the class predictor determines the final label based on these concept
predictions. The concept predictor is typically implemented using a
convolutional neural network (CNN), which extracts a feature vector encoding
the concepts from the input image. The class predictor, often implemented as a
downstream layer (usually a linear fully-connected layer), predicts the final
class using only the predicted concepts. This downstream layer serves as a
tool for interpreting the final class prediction in terms of the underlying
concepts \cite{chauhan2023interactive,Sarkar_2022_CVPR}. Additionally, the CBM
architecture allows for interactive intervention during the interpretation and
prediction process \cite{chauhan2023interactive}. The CBM and its numerous
modifications, such as \cite{kim2023probabilistic, ismail2023concept,
kazmierczak2024clipqda, marconato2022glancenets, vandenhirtz2024stochastic,
zarlenga2023tabcbm}, are considered some of the most notable representatives
of CBL models.

An important type of data that arises in many fields, including medicine,
reliability, safety, and economics, is event time data. Event time
observations can be approached within the framework of conventional regression
problems. However, unlike standard observations, some events may remain
unobserved because they occur after a fixed time point. Such data is referred
to as censored observations. Survival analysis \cite{Hosmer-Lemeshow-May-2008}
is a framework designed to handle two types of event time data: censored
(where the event of interest is not observed) and uncensored (where the event
of interest is observed). Unlike many traditional machine learning models, the
predictions of survival models are typically expressed as probabilistic
functions of time. For example, the survival function represents the
probability of an event not occurring up to a predefined time point.

It is worth noting that Forest et al. \cite{forest2024interpretable}
introduced a concept bottleneck model (CBM) for predicting the remaining
useful life. In their work, concepts are represented by different degradation
modes associated with the remaining useful life. However, this approach
remains within the standard regression framework of CBMs and does not address
censored data or incorporate survival analysis. To the best of our knowledge,
no existing method combines concept-based learning (CBL) with survival
analysis. Given the importance of tasks involving interpretable predictions
and improved prediction accuracy, we propose a novel approach that integrates
survival analysis into the CBM framework. This model, called \emph{SurvCBM}
(\textbf{Surv}ival \textbf{C}oncept-based \textbf{B}ottleneck \textbf{M}odel),
is designed to address survival analysis tasks.

The first idea underlying the proposed approach is to implement the second
predictor in the CBM as the Cox proportional hazards model \cite{Cox-1972},
where concepts serve as covariates with a linear relationship between them. In
this setup, the coefficients of the concept linear combination in the Cox
model, along with the covariates, can be interpreted as measures of the
concepts' impact on the predictions. These predictions are expressed in terms
of survival functions or other probabilistic measures within the survival
analysis framework \cite{Kovalev-Utkin-Kasimov-20a}. Alternatively, the second
predictor can be implemented using the Beran estimator \cite{Beran-81}, which
also provides a means to interpret predictions
\cite{Utkin-Eremenko-Konstantinov-24}. When using the Beran estimator, we can
adopt an approach within the framework of example-based explanations. This
approach involves selecting several instances from the training set that are
closest to the explainable instance, based on the proximity of their predicted
survival functions as measured by a distance metric. The importance of
concepts in the prediction is then determined by the number of matching
concepts between the closest instances and the explainable instance. The more
concepts that coincide, the greater their significance in the prediction. This
explanation approach is universal and can be applied to many survival models
integrated into SurvCBM for predicting the probability distributions of event times.

The second idea behind the proposed approach involves two distinct
architectures for CBL models. The primary architecture is a CBM with a
bottleneck layer consisting of concept logits, followed by either the Beran or
Cox models. The secondary architecture, introduced for comparison purposes,
also employs the Beran or Cox models but uses concepts as regularization to
enhance prediction accuracy. This architecture is referred to as
\emph{SurvRCM} (\textbf{Surv}ival \textbf{R}egularized \textbf{C}oncept-based
\textbf{M}odel). Both neural network architectures are trained in an
end-to-end manner.

Our contributions can be summarized as follows:

\begin{enumerate}
\item New concept-based survival models, SurvCBM and SurvRCM, are proposed for
dealing with censored data in the framework of survival analysis. These models
not only improve the accuracy of predictions but also provide a tool for
interpreting these predictions in terms of concepts. The models are based on
the Cox proportional hazards model and the Beran estimator.

\item Two approaches for interpreting the predicted survival functions are
considered, depending on the survival model used in SurvCBM or SurvRCM. The
first approach is based on the assumption of a linear relationship between
covariates in the Cox model. Using the trained regression coefficients and the
logits of the concepts corresponding to the explainable instance, the
importance of each concept is determined by its contribution to the linear
combination of covariates. The second approach is based on comparing instances
and their concepts, focusing on those with survival functions closest to the
survival function of the new explainable instance. The importance of concepts
is determined by the frequency of matching concepts between the closest
instances and the explainable instance.

\item The proposed models are compared with each other and with a survival
model that does not incorporate concepts.

\item Various numerical experiments are conducted to compare the proposed
concept-based survival models under different conditions. The results
demonstrate that the first architecture, SurvCBM, outperforms the other
models. Moreover, they demonstrate the importance and advantage of applying
concept information. The corresponding codes implementing the proposed models
are publicly available at: \url{https://github.com/NTAILab/SurvCBM}.
\end{enumerate}

The paper is organized as follows. Related work considering the existing
explanation methods can be found in Section 2. A short description of basic
concepts of survival analysis, including the Cox model and the Beran
estimator, as well as concept-based learning is given in Section 3. A general
idea of the concept-based survival models is provided in Section 4. Numerical
experiments comparing different architectures of the survival models are given
in Section 5. Approaches to interpreting predictions of concept-based survival
models are studied in Section 6. Concluding remarks are provided in Section 7.

\section{Related work}

\textbf{Concept-based learning models.} Many CBL models were proposed in
recent years \cite{kim2018interpretability,Sheth-Kahou-23,yeh2020completeness}
in order to improve the classification and regression performance of machine
learning models and to interpret their predictions in terms of the high-level
concepts. A large part of the models are CBMs \cite{koh2020concept} which have
attracted special attention due to a number of remarkable properties. In
particular, we point out stochastic CBMs \cite{vandenhirtz2024stochastic},
interactive CBMs \cite{chauhan2023interactive}, editable CBMs
\cite{hu2024editable}, semi-supervised CBMs \cite{hu2024semi}, probabilistic
CBMs \cite{kim2023probabilistic}, label-free CBMs \cite{oikarinen2023label},
CBMs without predefined concepts, \cite{Wang-Junlin-Chen-24}
\cite{schrodi2024concept}, post-hoc CBMs \cite{yuksekgonul2022post},
incremental residual CBMs \cite{shang2024incremental}, concept bottleneck
generative models \cite{ismail2023concept}, any CBMs
\cite{dominici2024anycbms}. concept complement bottleneck models
\cite{Wang-Junlin-Chen-24}. The above CBMs are a small part of various CBL
models proposed in literature.

Concept-based models, their advantages and disadvantages are considered in the
survey papers
\cite{Gupta-Narayanan-24,poeta2023concept,Aysel-etal-25,lee2023neural,mahinpei2021promises}%
.

\textbf{Survival analysis in machine learning}. Many survival machine learning
models have been developed \cite{Wang-Li-Reddy-2019} due to their importance
in several application areas, for example, in medicine, safety, reliability,
economics. Detailed reviews of many survival models can be found in
\cite{Wang-Li-Reddy-2019,salerno2023high,Wiegrebe:2024aa}. Deep survival
machine learning models were reviewed in \cite{chen2024introduction}. A
practical introduction to survival analysis was provided by Emmert-Streib and
Dehmer in \cite{EmmertStreib-Dehmer-19}.

A large part of survival models can be regarded as extensions of conventional
machine learning models under condition of censored data. For example, several
survival models are based on applying neural networks and deep learning
\cite{chen2024introduction,Katzman-etal-2018,Luck-etal-2017,Nezhad-etal-2018,ren2019deep,Steingrimsson-Morrison-20,Tarkhan-etal-21,Yao-Zhu-Zhu-Huang-2017,Zhong-Mueller-Wang-21}%
, several survival models are based on the transformer architectures
\cite{Chatha-etal-22,hu2021transformer,Li-Zhu-Yao-Huang-22,Lv-Lin-etal-22,Shen-liu-etal-22,tang2023explainable,Wang-Sun-22}%
, attention-based deep survival models were proposed in
\cite{Li-Krivtsov-Arora-22,Sun-Dong-etal-21}. A part of models is based on
extending the random forest \cite{Ibrahim-etal-2008,Wright-etal-2017}.
Convolutional neural networks also used in survival analysis
\cite{Haarburger-etal-2018}. Many machine learning survival models extend the
Cox model \cite{Cox-1972}. They mainly relax or modify the linear relationship
assumption used in the Cox model
\cite{Widodo-Yang-2011,Witten-Tibshirani-2010}.

Despite the intensive development of survival models, their application to
concept-based learning is currently not reflected in the literature.
Therefore, this work can be considered as the first attempt to create a
survival concept-based model.

\section{Background}

\subsection{Formal problem statement of concept-based learning}

In the conventional supervised concept-based model setting, a training set
consists of triplets $(\mathbf{x}_{i},y_{i},\mathbf{c}_{i})$, $i=1,...,n$,
where $\mathbf{x}_{i}\in$ $\mathbb{R}^{D}$ or $\mathbf{x}_{i}\in$
$\mathbb{R}^{d_{1}\times d_{2}}$ is the input instance represented as a vector
or a matrix; $y_{i}\in \mathcal{Y}=\{1,...,s\}$ is the corresponding target
defining $s$-class classification task; $\mathbf{c}_{i}=(c_{i}^{(1)}%
,...,c_{i}^{(m)})\in \mathcal{C}$ is a vector of $m$ high-level concepts which
describe the $i$-th instance $\mathbf{x}_{i}$ with $c_{i}^{(j)}\in
\{0,...,k_{j}-1\}$; $k_{j}$ is the number of the $j$-th concept values.
Concepts are usually represented as a vector consisting of binary elements
such that the $j$-th unit element corresponds to the case when the $j$-th
concept is presented in the description of the $i$-th instance. The main task
of CBL is to construct a machine learning models (a classifier or a regressor)
$h:\mathbb{R}^{D}\rightarrow(\mathcal{C},\mathcal{Y})$ to predict concepts and
the target for a new input instance $\mathbf{x}$.

Another task is to explain the predicted target in terms of concepts. One of
the approaches for solving this task is the CBM proposed by Koh et al.
\cite{koh2020concept}, which consists of two parts: the first part explicitly
predicts concept labels from instances and implements the map $g:$
$\mathcal{X}\rightarrow \mathcal{C}$, the second part predicts the target class
using only predicted concepts and implements the map $f:\mathcal{C}%
\rightarrow \mathcal{Y}$. The vector of concepts $\mathbf{c}$ or their
probabilities (logits) predicted by the first part can be viewed as a
bottleneck. As a result, the target class $y$ of a new input instance
$\mathbf{x}$ is defined as $y=f(g(\mathbf{x}))$.

\subsection{Elements of survival analysis}

Instances in survival analysis are represented by a triplet $(\mathbf{x}%
_{i},\delta_{i},T_{i})$, where $\mathbf{x}_{i}^{\mathrm{T}}=(x_{i}%
^{(1)},...,x_{i}^{(d)})\in \mathbb{R}^{d}$ is a vector of $d$ features
characterizing the $i$-th instance whose time to an event of interest is
$T_{i}$; $\delta_{i}$ is the censoring indicator taking the value $1$ if the
event of interest is observed at time $T_{i}$ (uncensored observation), and
$0$ if the event is not observed (censored observation)
\cite{Hosmer-Lemeshow-May-2008}. For censored observations, it is only known
that the time to the event exceeds the duration of observation. Survival
analysis aims to estimate the time to the event $T$ for a new instance
$\mathbf{x}$ by using the training set $\mathcal{A}=\{(\mathbf{x}_{i}%
,\delta_{i},T_{i}),i=1,...,n\}$.

One of the important elements of survival analysis is the survival function
(SF) denoted as $S(t\mid \mathbf{x})$, which is the probability of surviving
beyond time $t$, that is $S(t\mid \mathbf{x})=\Pr \{T>t\mid \mathbf{x}\}$.
Another element is the cumulative hazard function (CHF) denoted as
$H(t\mid \mathbf{x})$ an expressed through the SF as follows:
\begin{equation}
S(t\mid \mathbf{x})=\exp \left(  -H(t\mid \mathbf{x})\right)  .
\end{equation}

Another element of survival analysis used for comparison of different survival
models is the C-index \cite{Harrell-etal-1982}. It estimates the probability
that event times of a pair of instances are correctly ranking. Let
$\mathcal{J}$ be a set of all pairs $(i,j)$ satisfying conditions $\delta
_{i}=1$ and $T_{i}<T_{j}$. The C-index is formally computed as
\cite{Uno-etal-11,Wang-Li-Reddy-2019}:%
\begin{equation}
C=\frac{\sum_{(i,j)\in \mathcal{J}}\mathbf{1}[\widehat{T}_{i}<\widehat{T}_{j}%
]}{\sum_{(i,j)\in \mathcal{J}}1},
\end{equation}
where $\widehat{T}_{i}$ and $\widehat{T}_{j}$ are expected (predicted) event
times obtained from the predicted SFs $S(t\mid \mathbf{x}_{i})$ and
$S(t\mid \mathbf{x}_{j})$.

A popular semi-parametric regression model for analysis of survival data is
the Cox proportional hazards model that calculates the effects of observed
covariates on the risk of an event occurring \cite{Cox-1972}. The model
assumes that the log-risk of an event of interest is a linear combination of
covariates or features. According to the Cox model, the CHF at time $t$ given
instance $\mathbf{x}$ is defined as:
\begin{equation}
H(t\mid \mathbf{x},\mathbf{b})=H_{0}(t)\exp \left(  \mathbf{b}^{\mathrm{T}%
}\mathbf{x}\right)  ,
\end{equation}
where $H_{0}(t)$ is the baseline CHF estimated by using the Nelson--Aalen
estimator \cite{Hosmer-Lemeshow-May-2008,EmmertStreib-Dehmer-19};
$\mathbf{b}^{\mathrm{T}}=(b_{1},...,b_{m})$ is a vector of the model
parameters in the form of the regression coefficients which can be found by
maximizing the partial likelihood function for the dataset $\mathcal{A}$.

The SF $S(t\mid \mathbf{x},\mathbf{b})$ in the framework of the Cox model is
computed as
\begin{equation}
S_{C}(t\mid \mathbf{x},\mathbf{b})=\left(  S_{0}(t)\right)  ^{\exp \left(
\mathbf{b}^{\mathrm{T}}\mathbf{x}\right)  }. \label{Cox_SF}%
\end{equation}

Here $S_{0}(t)$ is the baseline SF which is also determined by using the
Nelson--Aalen estimator. It is important to note that functions $H_{0}(t)$ and
$S_{0}(t)$ do not depend on $\mathbf{x}$ and $\mathbf{b}$.

Another important model which is used below is the Beran estimator
\cite{Beran-81} which estimates the SF on the basis of the dataset
$\mathcal{A}$ as follows:
\begin{equation}
S_{B}(t\mid \mathbf{x})=\prod_{t_{i}\leq t}\left \{  1-\frac{\alpha
(\mathbf{x},\mathbf{x}_{i})}{1-\sum_{j=1}^{i-1}\alpha(\mathbf{x}%
,\mathbf{x}_{j})}\right \}  ^{\delta_{i}}, \label{Beran_est}%
\end{equation}
where time moments $t_{1},...,t_{n}$ are ordered; the weight $\alpha
(\mathbf{x},\mathbf{x}_{i})$ conforms with relevance of the $i$-th instance
$\mathbf{x}_{i}$ to the vector $\mathbf{x}$ and can be defined by using
kernels $K(\mathbf{x},\mathbf{x}_{i})$ as%
\begin{equation}
\alpha(\mathbf{x},\mathbf{x}_{i})=\frac{K(\mathbf{x},\mathbf{x}_{i})}%
{\sum_{j=1}^{n}K(\mathbf{x},\mathbf{x}_{j})}. \label{Weight1}%
\end{equation}

In particular, if to use the Gaussian kernel with the parameter $\tau$, then
weights $\alpha(\mathbf{x},\mathbf{x}_{i})$ are of the form:
\begin{equation}
\alpha(\mathbf{x},\mathbf{x}_{i})=\text{\textrm{softmax}}\left(
-\frac{\left \Vert \mathbf{x}-\mathbf{x}_{i}\right \Vert ^{2}}{\tau}\right)  .
\label{Weight2}%
\end{equation}

The Beran estimator is reduced to the Kaplan-Meier estimator
\cite{Wang-Li-Reddy-2019} when all weights are identical, i.e., $\alpha
(\mathbf{x},\mathbf{x}_{i})=1/n$ for all $i=1,...,n$.

\section{Description of concept-based survival models}

\subsection{Concept-based survival problem statement}

In the survival concept-based learning setting, a training set $\mathcal{A}$
is represented by a set of $n$ quadruplets $(\mathbf{x}_{i},\mathbf{c}%
_{i},T_{i},\delta_{i})$, $i=1,...,n$, which combines concept-based and
survival information considered above. Unlike most concept-based learning
models, predictions in the survival CBL are SFs $S(t\mid \mathbf{x}%
,\mathbf{c})$ or CHFs $H(t\mid \mathbf{x},\mathbf{c})$. For simplicity, we will
consider only SFs because CHFs are expressed through the corresponding SFs.
The concept-based survival analysis aims to estimate the SF $S(t\mid
\mathbf{x})$ as well as the concept vector $\mathbf{c}$ for a new instance
$\mathbf{x}$ by using the training set $\mathcal{A}$. Other elements of
survival analysis, for example, the mean time to event, can be calculated on
the basis of the SF. If we denote the set of SFs as $\mathcal{S}$, then our
task is to construct machine learning models which implement the map
$G:\mathbb{R}^{D}\rightarrow(\mathcal{C},\mathcal{S})$. Two models are
proposed and compared in this work.

\subsection{The first model: SurvCBM}

The second proposed model is derived from the architecture of the CBM. Its
mathematical formulation can be expressed as follows:%

\begin{equation}
S(t\mid \mathbf{x},\mathcal{A})=F(G(\mathbf{x})),
\end{equation}
where $S(t\mid \mathbf{x},\mathcal{A})$ is the SF obtained for a feature vector
$\mathbf{x}$ using training set $\mathcal{A}$; $F(\mathbf{c})$ is a survival
model defined over the concept space; $G(\mathbf{x})$ is a neural network (or
a set of neural networks) that predicts the vector of concepts for the feature
vector $\mathbf{x}$.

\begin{figure}
[ptb]
\begin{center}
\includegraphics[
height=4.3431in,
width=4.6458in
]%
{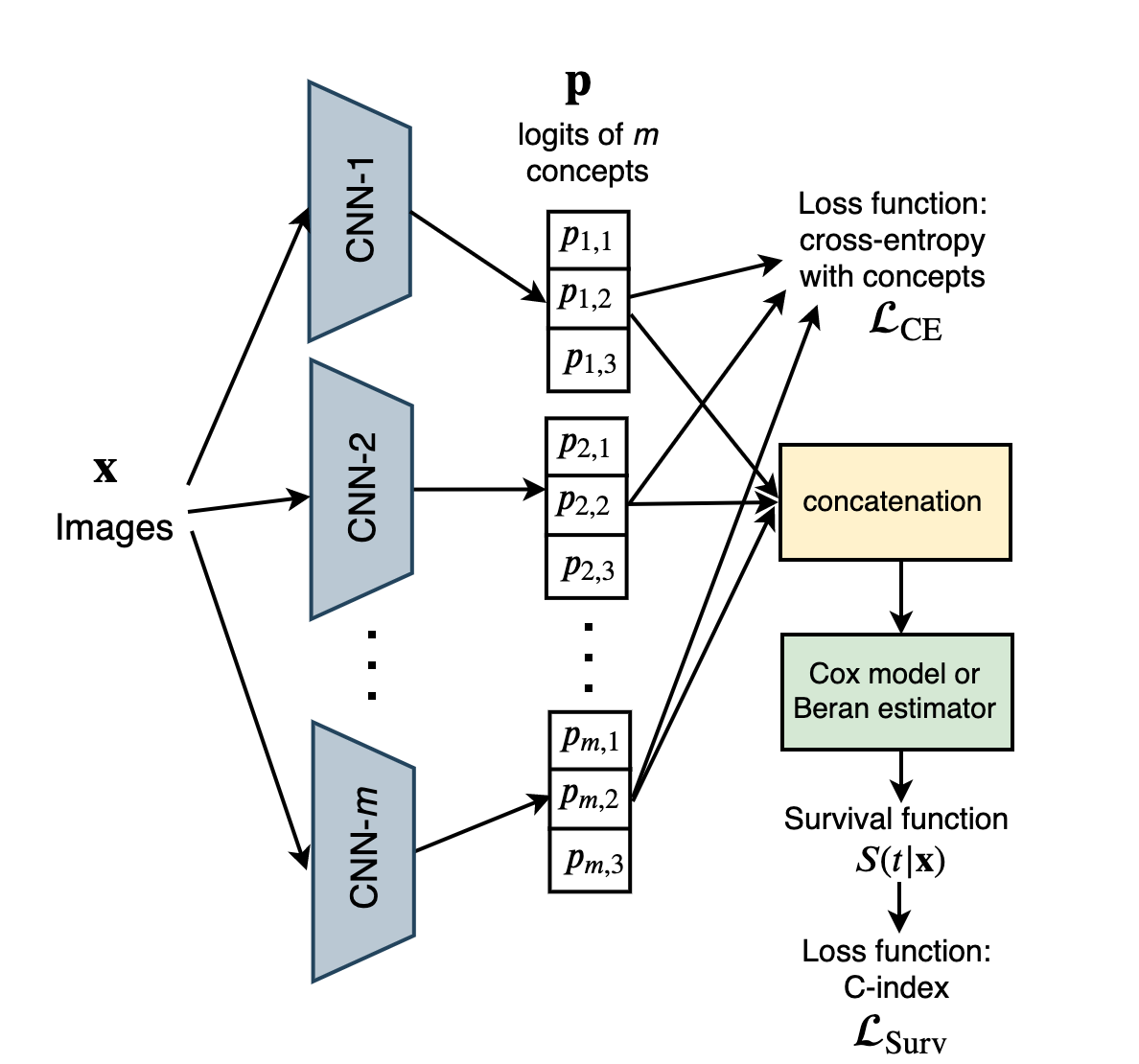}%
\caption{The architecture of SurvCBM}%
\label{f:surv_bottleneck}%
\end{center}
\end{figure}

The architecture of the model is illustrated in Fig. \ref{f:surv_bottleneck}.
It can be seen from Fig. \ref{f:surv_bottleneck} that the input image
$\mathbf{x}$ is processed by a set of $m$ convolutional neural networks
(CNN-1,...,CNN-$m$) where each network outputs a vector of concept logits
$\mathbf{p}_{i}=(p_{i,1},...,p_{i,k_{i}})$. Here, it is assumed that the
$i$-th concept can take $k_{i}$ possible values. The use of CNNs is crucial
for reducing the dimensionality of the images and enabling the training of Cox
or Beran models on lower-dimensional vectors. In SurvCBM, the Cox model and
the Beran estimator are trained on the concatenated logits
\begin{equation}
\mathbf{p}=(\mathbf{p}_{1},...,\mathbf{p}_{m})=(p_{i,1},...,p_{i,k_{i}%
},i=1,...,m)\in \mathbb{R}^{M},\  \ M=\sum_{i=1}^{m}k_{i}.
\end{equation}

This is a very important peculiarity of SurvCBM, as it achieves a balance
between two objectives: accurately defining the concept values and computing
the SF. Furthermore, this design ensures that the model remains interpretable,
and the computed SF can be explained in terms of the underlying concepts.

The general form of the loss function is given by:%
\begin{equation}
\mathcal{L}=-\alpha \mathcal{L}_{\text{surv}}+(1-\alpha)\mathcal{L}_{\text{CE}%
}, \label{gen_loss}%
\end{equation}
where $\alpha \in(0,1)$ is a hyperparameter.

The term $\mathcal{L}_{\text{CE}}$ corresponds to solving the concept
classification task. For a single data point $\mathbf{x}$ with true concept
values $(y_{1},...,y_{m})$, it is defined as:
\begin{equation}
\mathcal{L}_{\text{CE}}=-\frac{1}{m}\sum \limits_{i=1}^{m}\log \frac
{\exp{(p_{i,y_{i}})}}{\sum_{j=1}^{k_{i}}\exp(p_{i,j})}. \label{loss_entropy}%
\end{equation}

The term $\mathcal{L}_{\text{surv}}$ corresponds to solving the survival
analysis task. To implement We apply $\mathcal{L}_{\text{surv}}$, we use the
smoothed C-index function. For data points $\mathbf{x}_{1},...,\mathbf{x}_{n}$
with true event times $T_{1},...,T_{n}$, censoring indicators $\delta
_{1},...,\delta_{n}$ and estimated expected event times $\hat{T}_{1}%
,...,\hat{T}_{n}$, the loss function can be expressed as:%

\begin{equation}
\mathcal{L}_{\text{surv}}=\frac{\sum_{i,j}\mathbb{I}_{T_{j}<T_{i}}\cdot
\sigma(\hat{T}_{i}-\hat{T}_{j})\cdot \delta_{j}}{\sum_{i,j}\mathbb{I}%
_{T_{j}<T_{i}}\cdot \delta_{j}}, \label{loss_C-index}%
\end{equation}
where $\sigma$ is the sigmoid function with a temperature parameter $\omega$,
which is treated as a hyperparameter.

The loss function $\mathcal{L}_{\text{CE}}$ is averaged over a batch during
the optimization process, while the smoothed C-index is computed over all
instances in the batch.

SurvCBM is trained in an end-to-end manner. It is worth noting that the use of
a set of convolutional neural networks is one possible implementation of the
model. Alternatively, a single network could be used to compute the vector
$\mathbf{p}$ of logits. However, our numerical experiments have demonstrated
that using a set of networks yields more robust results.%

\subsection{The second model: SurvRCM}

\begin{figure}
[ptb]
\begin{center}
\includegraphics[
height=3.9842in,
width=5.0747in
]%
{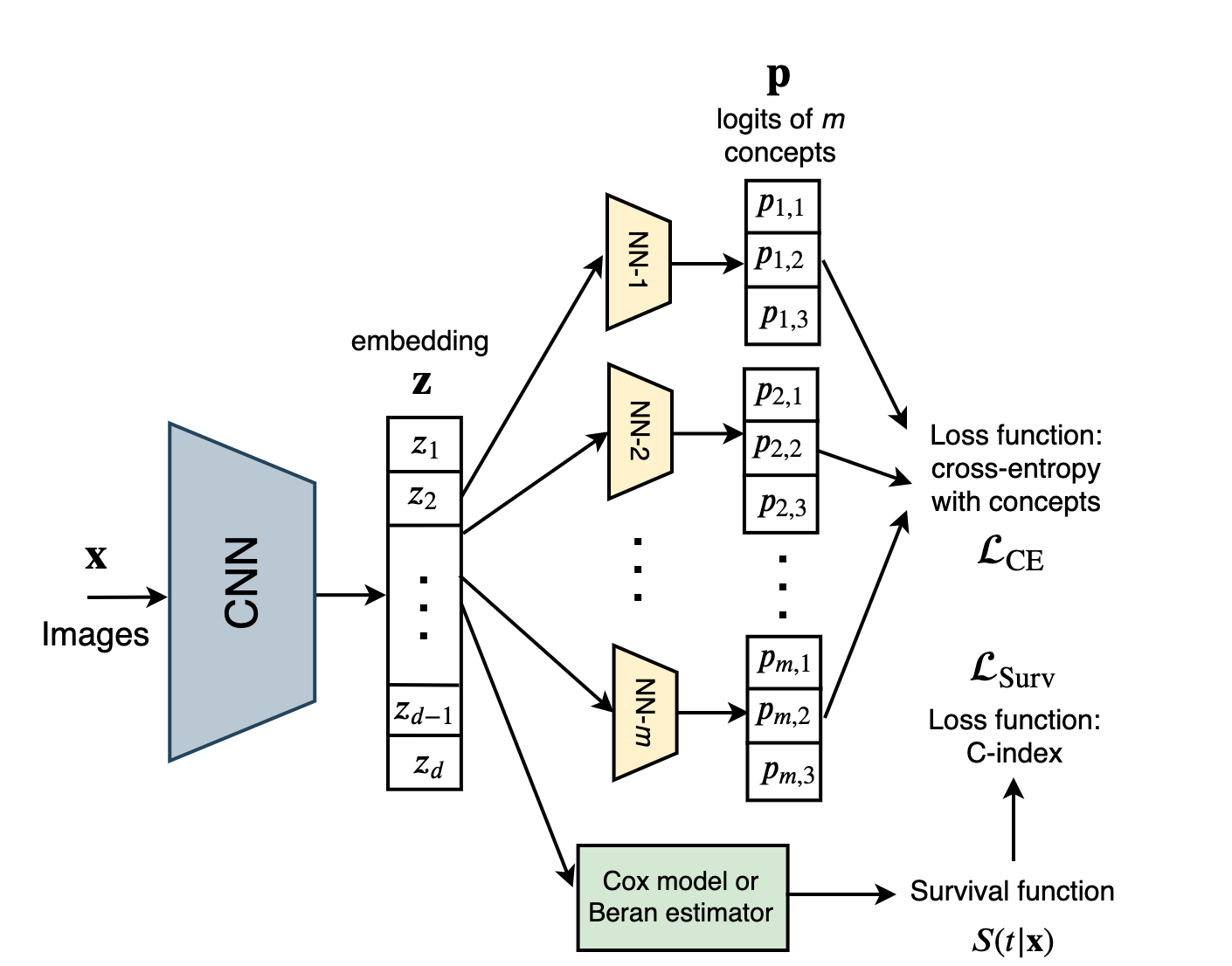}%
\caption{The architecture of SurvRCM}%
\label{f:surv_mixture}%
\end{center}
\end{figure}

The SurvRCM does not follow the architecture of the CBM because it lacks a
bottleneck layer. Nevertheless, we consider SurvRCM as a simplified variant of
concept-based survival models. Its architecture is illustrated in Fig.
\ref{f:surv_mixture}. It can be seen from Fig. \ref{f:surv_mixture} that
SurvRCM addresses the survival analysis task using concept information but
without the bottleneck scheme. In this model, a CNN generates an embedding
$\mathbf{z}\in \mathbb{R}^{d}$, which is fed to the Cox or Beran models to
produce the SF $S(t\mid \mathbf{x})$. The CNN is again used to reduce the
dimensionality of images and to learn the Cox or Beran models on
lower-dimensional vectors. The embedding $\mathbf{z}$ is also passed to a set
of fully connected neural networks (NN-1,...,NN-$m$) to obtain the logits for
$m$ concepts. Each network produces a vector of logits for the $i$-th concept
$\mathbf{p}_{i}=(p_{i,1},...,p_{i,k_{i}})$, $i=1,...,m$. Consequently, the Cox
model or the Beran estimator are trained on the embeddings $\mathbf{z}$, but
concepts are trained on logits $\mathbf{p}$.

The loss function for training SurvRCM is identical to that of SurvCBM, as
given in (\ref{gen_loss}). However, unlike SurvCBM, the term $\mathcal{L}%
_{\text{surv}}$ is computed under condition that the survival models (the Cox
and Beran models) are trained on embeddings $\mathbf{z}$, but not on the
vector of logits $\mathbf{p}$ as in SurvCBM. The model is also trained in an
end-to-end manner.%

The main drawback of the SurvRCM model is that the loss function
$\mathcal{L}_{\text{CE}}$ acts as a regularization term. It constrains the
logits of the concepts $\mathbf{p}$, but does not directly influence the
embeddings $\mathbf{z}$. As a result, the SF may be learned inaccurately, as
it is derived from the embeddings rather than the concept logits. This
limitation prevents the model from being fully interpretable, as the obtained
concepts do not directly explain the SF.

\section{Numerical Experiments}

Two types of numerical experiments are conducted. The first type evaluates the
performance of the proposed models, while the second type demonstrates their
explanation mechanisms.

\begin{figure}
[ptb]
\begin{center}
\includegraphics[
height=2.0539in,
width=4.0326in
]%
{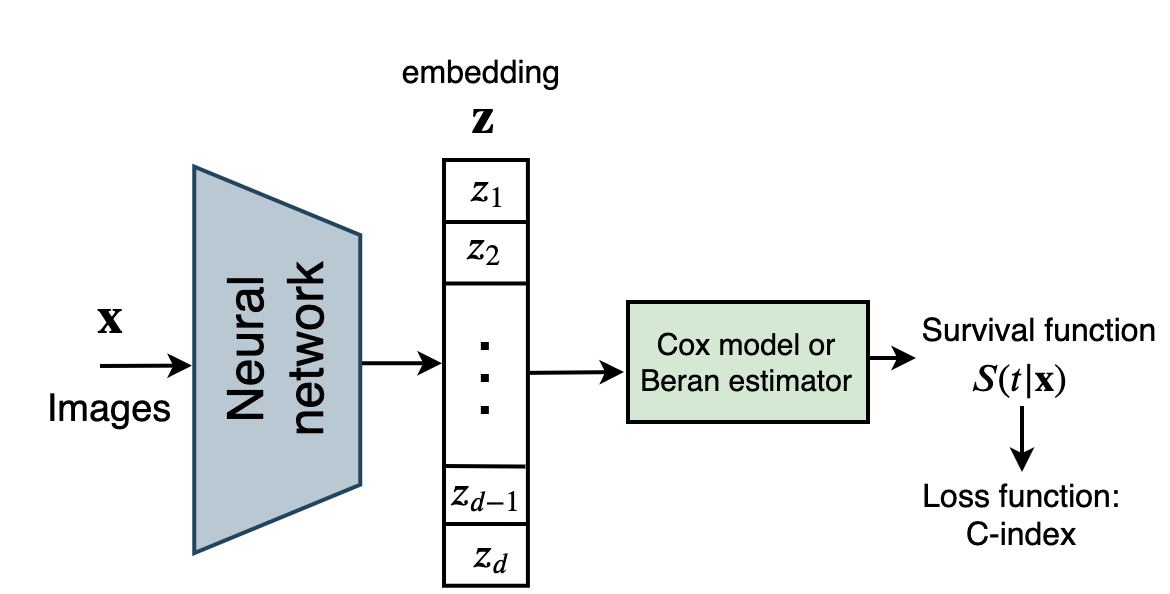}%
\caption{An architecture of SurvBase }%
\label{f:surv_baseline}%
\end{center}
\end{figure}

To assess the model performance, we compare the proposed models with each
other and with a baseline survival model. Due to the absence of known survival
CBMs, we compare the proposed models with a baseline model that solves the
same survival analysis task without incorporating concept data. This
comparison allows us to study the importance of correctly selected concepts
for solving survival tasks when the input instances are images. In the
baseline model, denoted as \emph{SurvBase}, a CNN generates an embedding
$\mathbf{z}$, which is used to train either the Cox model or the Beran
estimator. The model is trained in an end-to-end manner, and its loss function
consists solely of the term $\mathcal{L}_{\text{surv}}$, as defined in
(\ref{loss_C-index}). The architecture of SurvBase is illustrated in Fig.
\ref{f:surv_baseline}. It can seen from Fig. \ref{f:surv_baseline} that an
image $\mathbf{x}$ is fed into the CNN to produce an embedding $\mathbf{z}%
\in \mathbb{R}^{d}$. The Cox or Beran model is then applied to the set of
embeddings to generate the SF $S(t\mid \mathbf{x})$.%

The following model components and hyperparameters are used: the temperature
$\omega$ of the sigmoid function in the smoothed C-index; parameter $\alpha$
in the loss function; the number of instances in the background set of the
Beran estimator; the parameter $\tau$ of the Gaussian kernel in the Beran
estimator; the number of tasks, generated on each training epoch; the number
of epochs; the optimizer and its parameters, including learning rate and
weight decay; the size $d$ of the embedding $\mathbf{z}$. Different values of
the hyperparameters are tested, choosing those leading to the best results.

Experiments are conducted in two directions: comparison of the model
performance depending on the number of training instances and depending on the
number of uncensored data. The performance measures for comparison of the
models are C-index, characterizing the survival model, and F1-measure,
characterizing the concept classification accuracy. At that, values of the
F1-measures indicated on graphs below are averaged over all concepts for brevity.

The cross-validation in all experiments is performed with 100 repetitions.
Intervals on the graphs are the average value of these repetitions and the
standard deviation.

\subsection{Datasets}

\begin{figure}
[ptb]
\begin{center}
\includegraphics[
height=3.2785in,
width=3.0865in
]%
{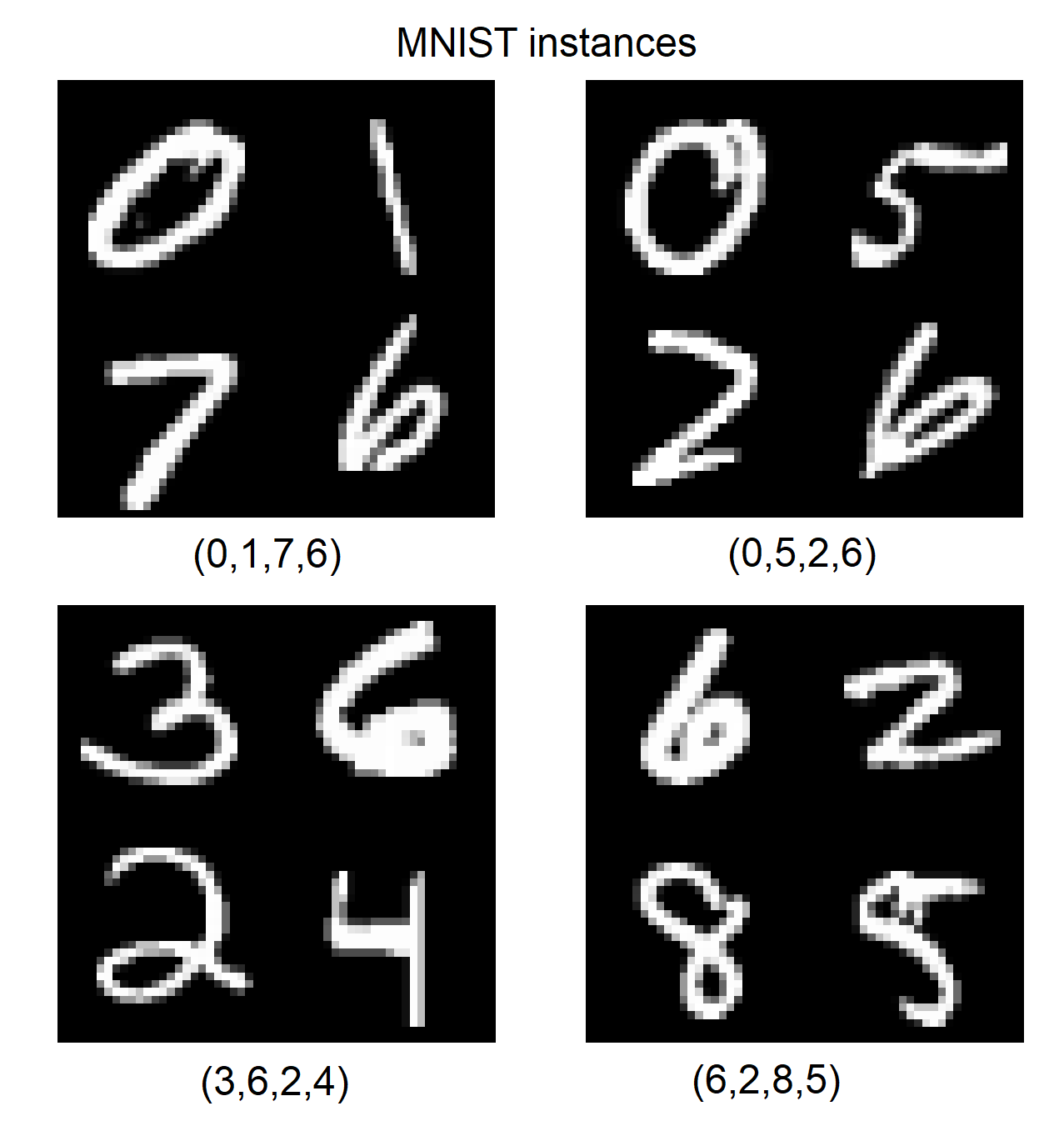}%
\caption{Instances constructed from the MNIST dataset; the corresponding
concepts are denoted under the pictures in accordance with digits}%
\label{f:mnist_example}%
\end{center}
\end{figure}

\begin{figure}
[ptb]
\begin{center}
\includegraphics[
height=3.1332in,
width=3.1332in
]%
{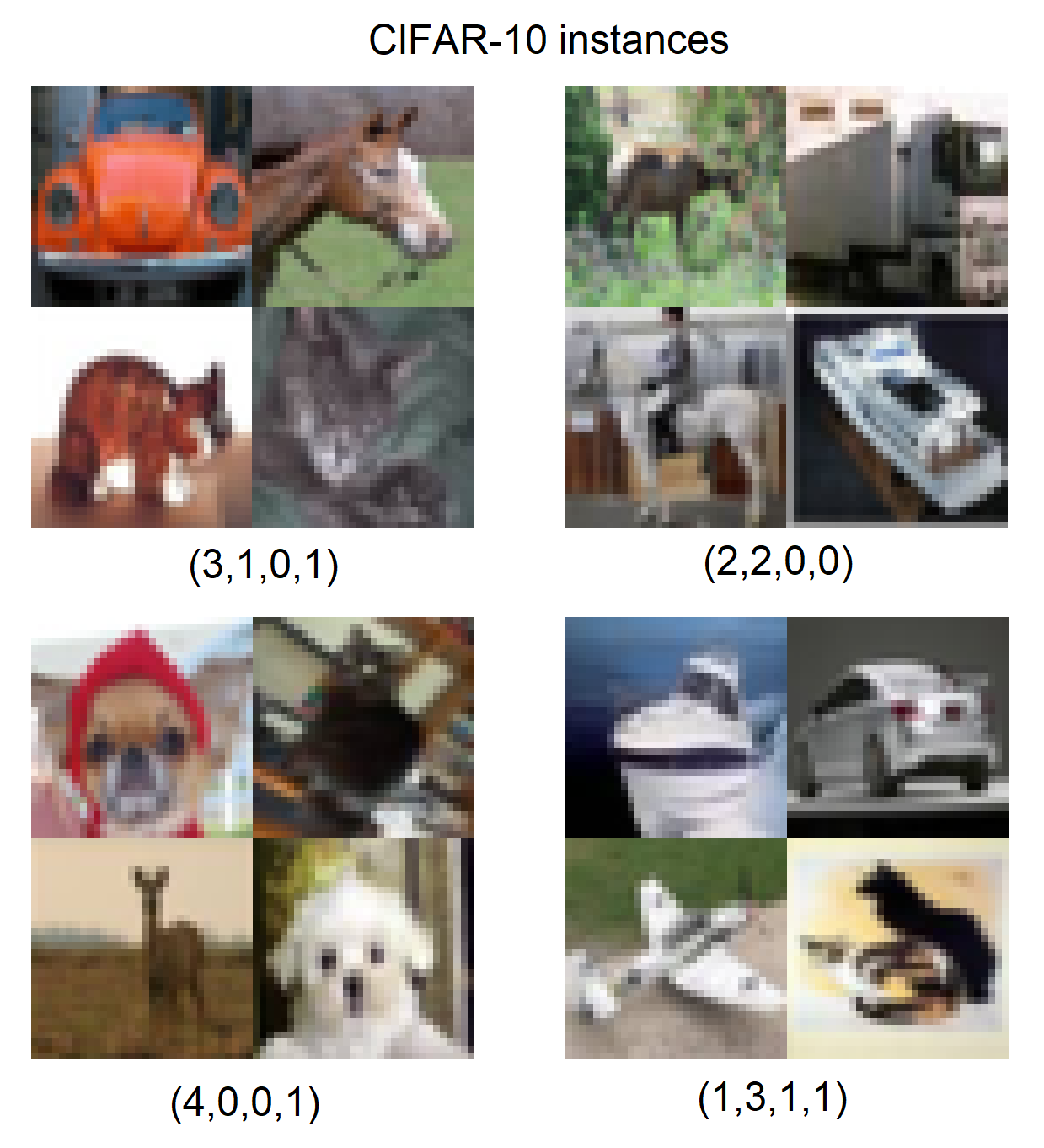}%
\caption{Instances constructed from the CIFAR-10 dataset; the corresponding
concepts are denoted under the pictures in accordance with their meanings}%
\label{f:cifar_example}%
\end{center}
\end{figure}

We consider the following composite images constructed from the datasets MNIST
and CIFAR-10 with synthetically generated concepts and target event times:

\begin{enumerate}
\item \textbf{MNIST}. First, we apply MNIST dataset which is a commonly used
large dataset of $28\times28$ pixel handwritten digit images
\cite{LeCun-etal-98}. It has a training set of 60,000 instances, and a test
set of 10,000 instances. The dataset is available at
http://yann.lecun.com/exdb/mnist/. To construct the concept-based MNIST
dataset, images from the original MNIST dataset are combined into sets of four
different digits. Each digit in its position defines a separate concept, where
the concept's value corresponds to the digit itself. As a result, each data
point in the dataset is described by four concepts, each taking one of 10
possible values. The event time is generated according to the Weibull
distribution and is computed as follows:
\begin{equation}
T=\left(  -\frac{\ln(u)}{\lambda \exp \left(  \mathbf{b}^{\mathrm{T}}%
\mathbf{c}\right)  }\right)  ^{\frac{1}{\nu}}, \label{Weibull}%
\end{equation}
where $u\sim U(0,1)$ is the uniform random variable, $\mathbf{c}$ is the
vector of concepts; $\nu$ and $\lambda$ are parameters of the Weibull distribution.

The generation parameters are $\mathbf{b}^{\mathrm{T}}=(0.5,1.5,-1,0.001)$%
,\ $\nu=2$, $\lambda=10^{-4}$. Values of the censoring indicator $\delta$ are
generated in accordance with the Bernoulli distribution. Examples of the
generated instances as well as vectors of concepts $\mathbf{c}$ are shown in
Fig. \ref{f:mnist_example}.

\item \textbf{MNIST-sin}. The second dataset is identical to the first one,
but the event times are generated using a probability distribution different
from the Weibull distribution. The event times are generated according to the
following formula:%
\begin{equation}
T=\left(  -\frac{\ln(u)}{\lambda \left(  \sin \left(  \mathbf{b}^{\mathrm{T}%
}\mathbf{c}\right)  +1.001\right)  }\right)  ^{\frac{1}{\nu}},
\label{Weibull-sin}%
\end{equation}
where the generation parameters are $\mathbf{b}^{\mathrm{T}}%
=(0.5,1.5,-1,0.001)$,\ $\nu=4$, $\lambda=0.01$.

\item \textbf{CIFAR-10}. The third dataset consists of four images from the
well-knwon CIFAR-10 dataset \cite{Krizhevsky-Hinton-2009}, which contains
$60000$ color images $32\times32$ drawn from 10 categories (50,000 training
and 10,000 test images each). The dataset is available at
https://www.cs.toronto.edu/\symbol{126}kriz/cifar.html. The following concepts
vector of instances consisting of four CIFAR-10 images is used: (1 - number of
animals, 2 - number of vehicles, 3 - number of flying objects, 4 - is there a
cat on the picture). Concepts 1, 2, 3 take five values, the last one is
binary. The event time is generated according to the Weibull distribution in
the same way as for the MNIST dataset. The generation parameters are
$\mathbf{b}=(-0.7,1.5,-2,5)$,\ $\nu=2$, $\lambda=0.01$. Examples of the
dataset can be found in Fig. \ref{f:cifar_example}. As an example, it can be
seen from the first instance in Fig. \ref{f:cifar_example} that it contains
three animals, one vehicle, no flying objects, and there is a cat. This
implies that the corresponding vector of concepts is $(3,1,0,1)$.
\end{enumerate}

%

%

\subsection{MNIST dataset}

First, we conduct numerical experiments using a synthetic dataset derived from
the MNIST dataset. All metrics presented in the graphs are computed on the
test set, which comprises 40\% of the instances from the dataset. For
experiments involving varying sample sizes, we set the proportion of
uncensored instances to 33\%.

\begin{figure}
[ptb]
\begin{center}
\includegraphics[
height=2.4431in,
width=4.8732in
]%
{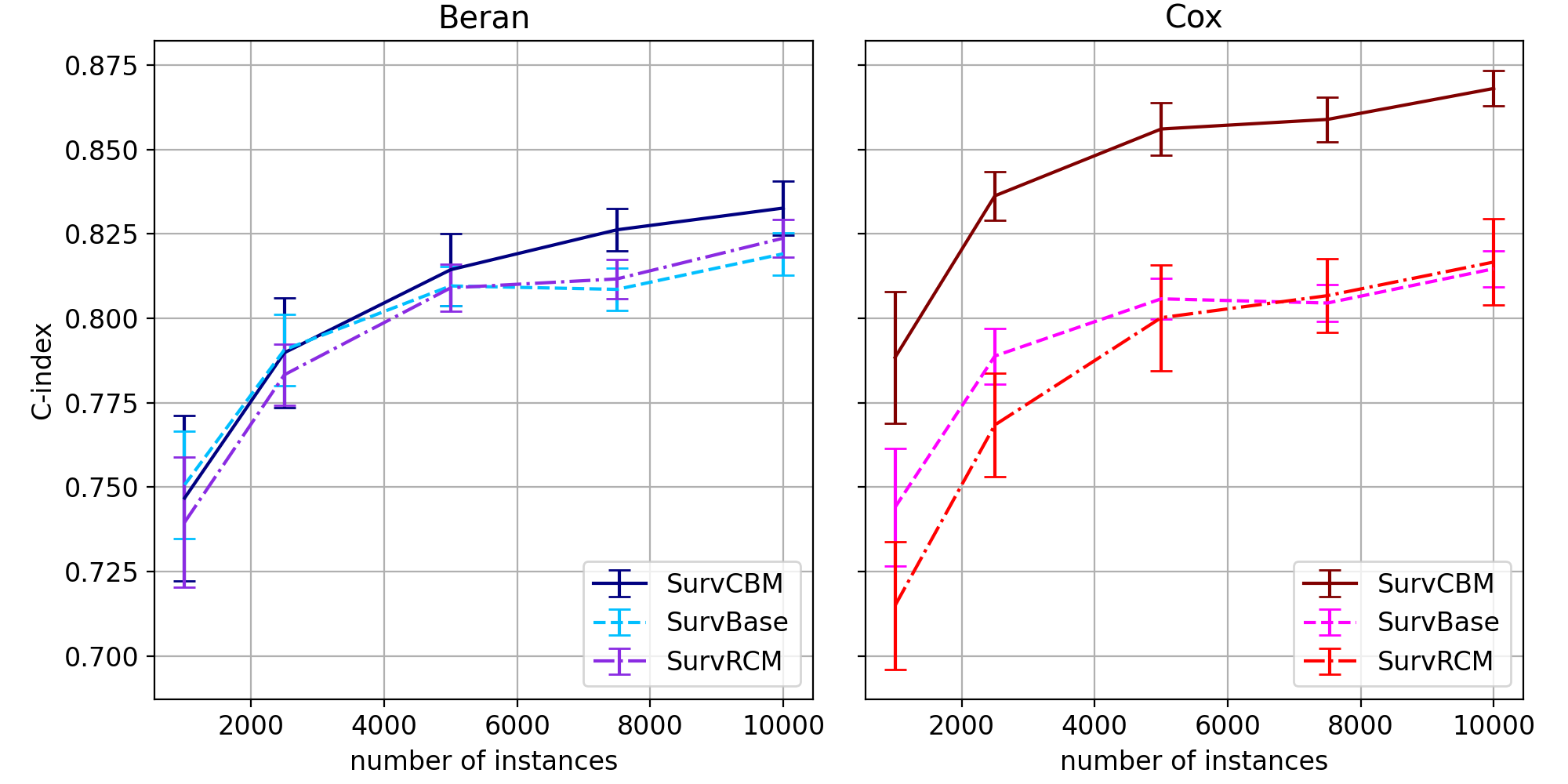}%
\caption{Dependencies of the survival model performance (the C-index) on the
number of training instances for the Beran estimator (the left graph) and the
Cox model (the right graph) when the MNIST dataset is used}%
\label{fig:N_MNIST_C_IND}%
\end{center}
\end{figure}

First, we study how the sample size $n$ impacts the accuracy (the C-index) of
the models. Fig. \ref{fig:N_MNIST_C_IND} illustrates how the C-index of the
compared models depends on the number of training instances $n$, when the
Beran estimator (the left graph) and the Cox model (the right graph) are used
as the survival models. It can be seen from the both graphs in Fig.
\ref{fig:N_MNIST_C_IND} that SurvCBM outperforms the other models. It is
interesting to point out that the most impressive results are demonstrated
when the Cox model is used. This is due to the relatively simple feature space
and the Weibull distribution used for generating the event times. It is also
interesting to note that SurvRCM provides worse results even in comparison
with SurvBase. We can also observe that the C-index of SurvBase is much
smaller than that of SurvCBM. This implies that the concept information may
significantly impove the model performance.%

\begin{figure}
[ptb]
\begin{center}
\includegraphics[
height=2.4474in,
width=4.8819in
]%
{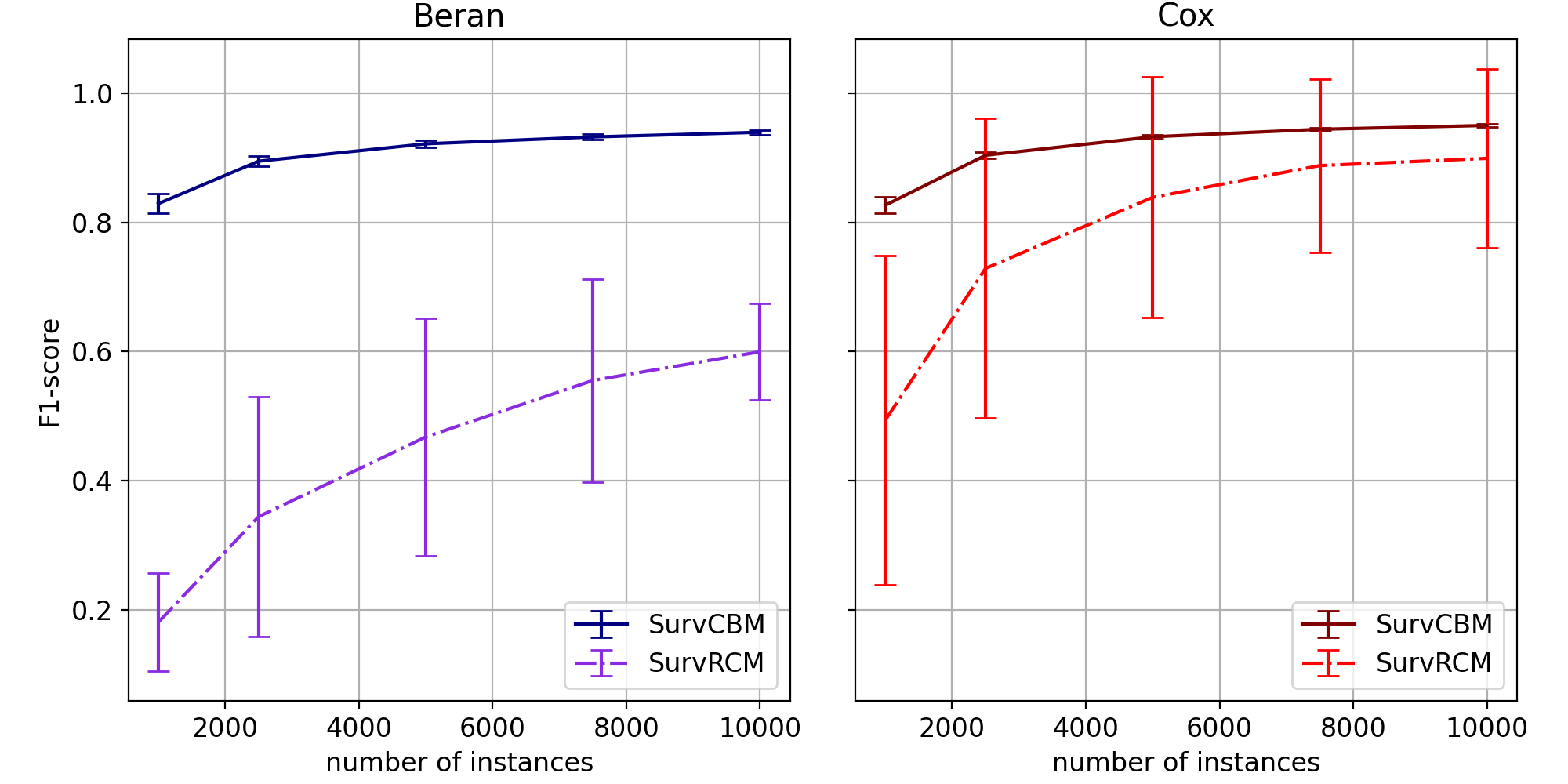}%
\caption{Dependencies of the concept classification performance (the
F1-measure) on the number of training instances for the Beran estimator (the
left graph) and the Cox model (the right graph) when the MNIST dataset is
used}%
\label{fig:N_MNIST_F1}%
\end{center}
\end{figure}

If the C-index characterizes the performance of models as survival models,
then F1-measure characterizes the concept classification accuracy. Fig.
\ref{fig:N_MNIST_F1} shows how the F1-measures of the models depend on the
number of training instances under the same conditions. We consider only two
models because SurvBase does not deal with concepts. It can be seen from Fig.
\ref{fig:N_MNIST_F1} that SurvCBM significanlty outperforms the Mixture model.
Moreover, results for SurvCBM for different survival models (the Beran and Cox
models) are almost the same. It is important to note that the variance of
SurvCBM results is extremely small compared with the variance of SurvRCM
results. This fact illustrates a high robustness of SurvCBM.%

\begin{figure}
[ptb]
\begin{center}
\includegraphics[
height=2.4742in,
width=4.9338in
]%
{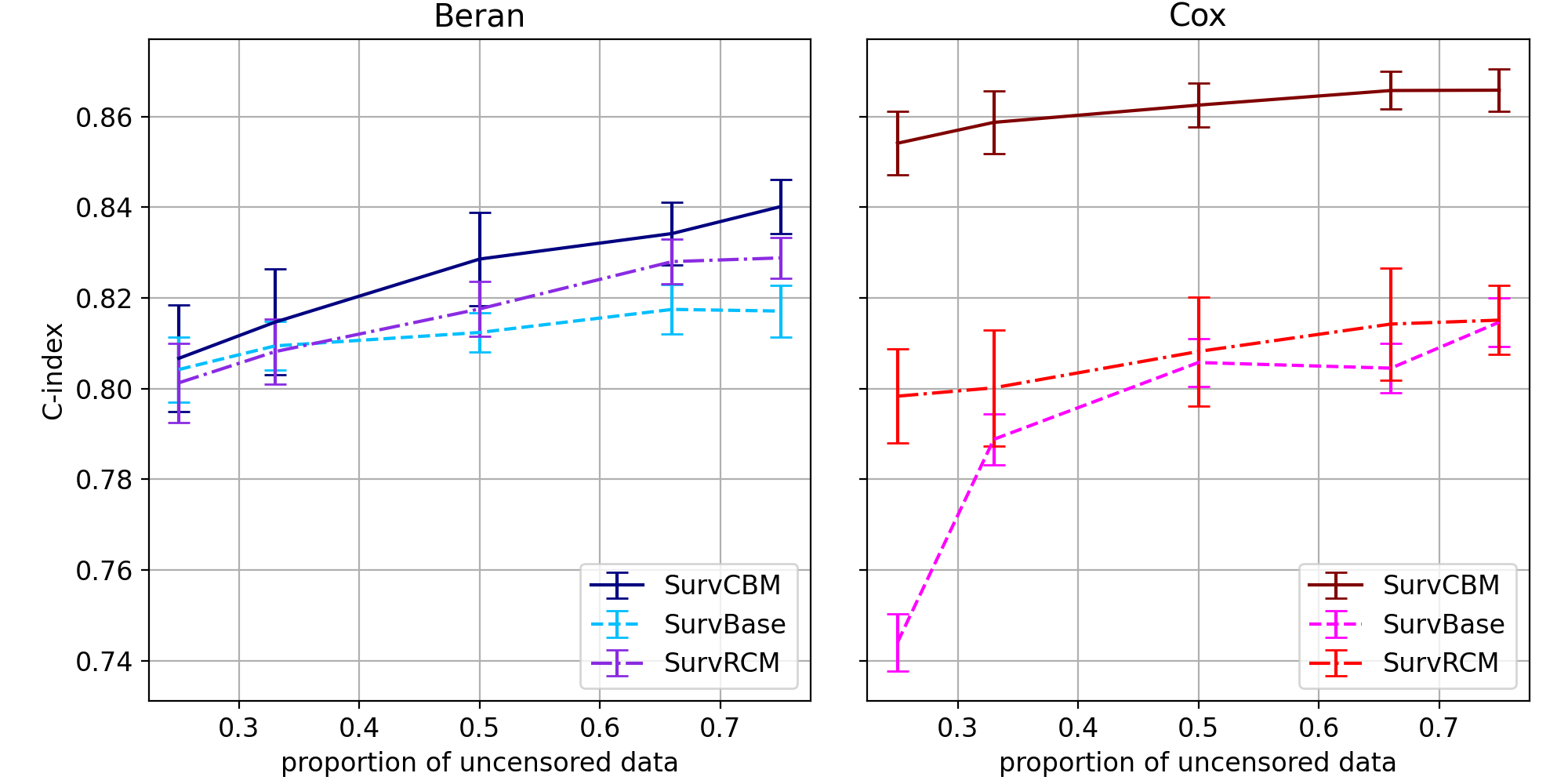}%
\caption{Dependencies of the survival model performance (the C-index) on the
proportion of uncensored data for the Beran estimator (the left graph) and the
Cox model (the right graph) when the MNIST dataset is used}%
\label{f:mnist_c_c_ind}%
\end{center}
\end{figure}

\begin{figure}
[ptb]
\begin{center}
\includegraphics[
height=2.4526in,
width=4.8905in
]%
{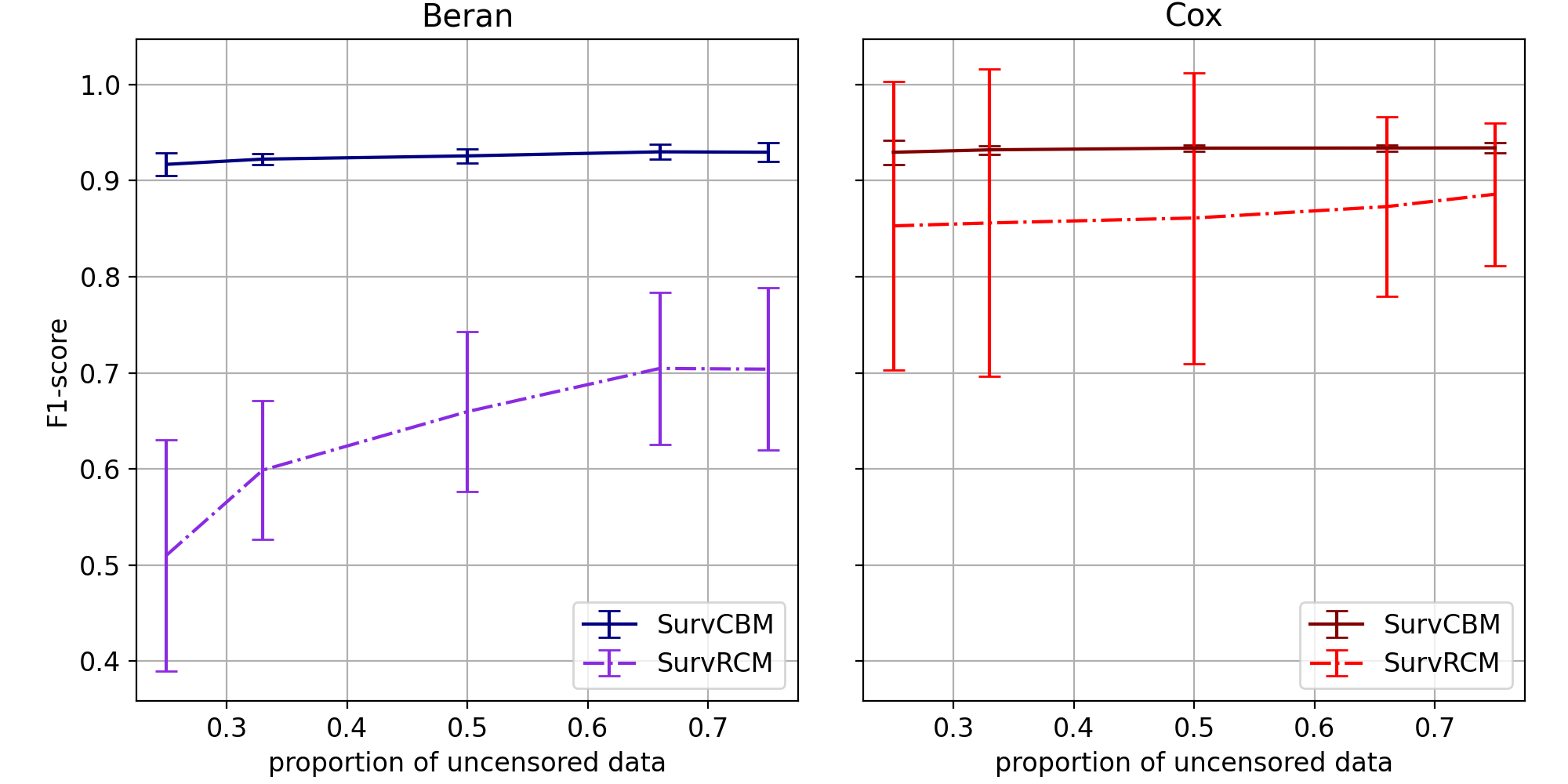}%
\caption{Dependencies of the survival model performance (the F1-measure) on
the proportion of uncensored data for the Beran estimator (the left graph) and
the Cox model (the right graph) when the MNIST dataset is used}%
\label{f:mnist_c_f1}%
\end{center}
\end{figure}

The next question is how the model performance depends on the proportion of
uncensored data $\rho$. Figs. \ref{f:mnist_c_c_ind} and \ref{f:mnist_c_f1}
address this question. In particular, Fig. \ref{f:mnist_c_c_ind} shows that
the C-index of SurvCBM significantly exceeds that of other models when the Cox
model is used as the survival model. This can be explained by the fact that
the Weibull distribution is used for generating the event times. One can also
see from Fig. \ref{f:mnist_c_c_ind} that SurvBase provides unsatisfactory
results. This fact again demonstrates that concept information may
significantly improve the model performance. Fig. \ref{f:mnist_c_f1} is
similar to Fig. \ref{fig:N_MNIST_F1}. The F1-measure of SurvCBM weakly depends
on the survival models, whereas SurvRCM strongly depends on both the survival
models and the parameter $\rho$.%

%

\subsubsection{MNIST-sin}

\begin{figure}
[ptb]
\begin{center}
\includegraphics[
height=2.5218in,
width=5.0298in
]%
{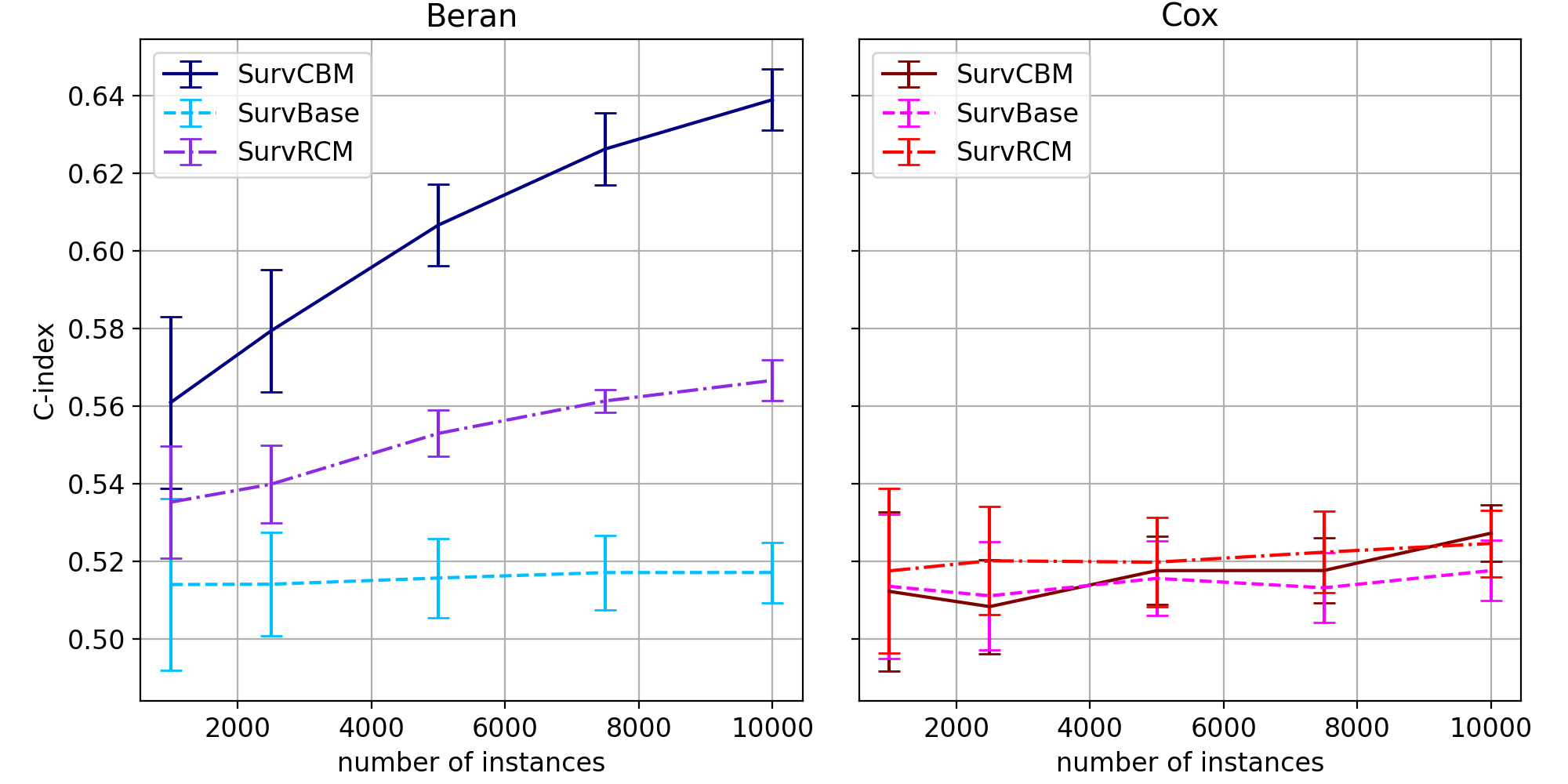}%
\caption{Dependencies of the survival model performance (the C-index) on the
number of training instances for the Beran estimator (the left graph) and the
Cox model (the right graph) when the MNIST-sin dataset is used}%
\label{f:N_MNIST_SIN_C_IND}%
\end{center}
\end{figure}

Let us study the model performance when the models are trained on the
MNIST-sin dataset. In this dataset, the Weibull distribution of the event
times is violated, and the proportionality of risk is also violated. The
consequences of these violation are clearly seen from Fig.
\ref{f:N_MNIST_SIN_C_IND} where the Beran estimator demonstrates superior
results compared to the Cox model when SurvCBM is used. This is because the
Beran model effectively handles complex data structures by taking into account
relationships between instances. It is interesting to note that all models
have similar C-indices when the Cox model is used. Moreover, increasing the
number of training instances does not significantly improve the performance of
any of the models.%

\begin{figure}
[ptb]
\begin{center}
\includegraphics[
height=2.5823in,
width=5.1508in
]%
{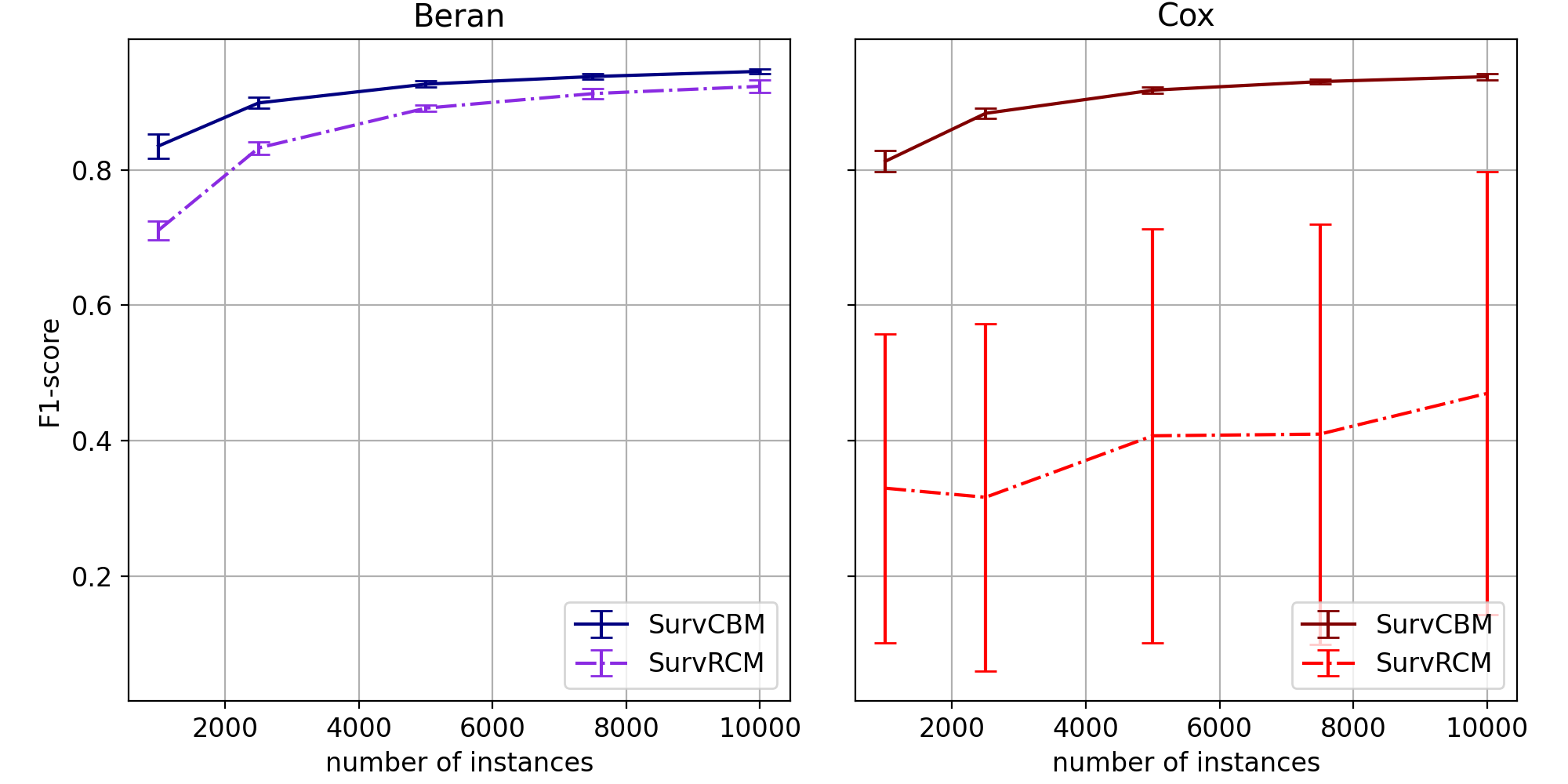}%
\caption{Dependencies of the survival model performance (the F1-measure) on
the number of training instances for the Beran estimator (the left graph) and
the Cox model (the right graph) when the MNIST-sin dataset is used}%
\label{f:mnistsin_n_f1}%
\end{center}
\end{figure}

Fig. \ref{f:mnistsin_n_f1} shows how the F1-measures of the models depend on
the number of training instances. It can be seen from Fig.
\ref{f:mnistsin_n_f1} that SurvCBM outperforms SurvRCM for both the Beran
estimator and the Cox model. We again observe that the F1-measures for both
models, using either the Beran or Cox models, are almost the same.%

\begin{figure}
[ptb]
\begin{center}
\includegraphics[
height=2.655in,
width=5.2987in
]%
{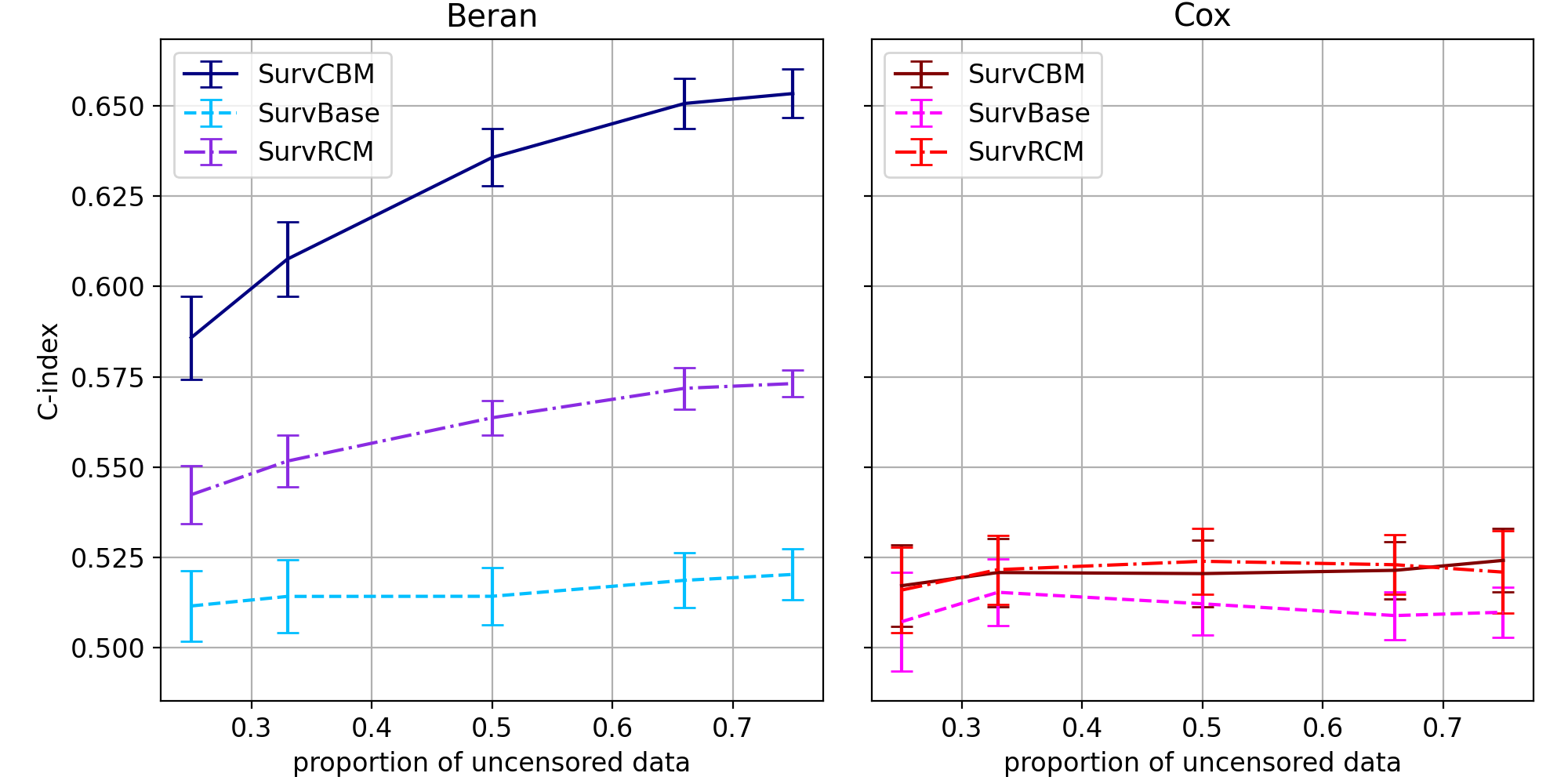}%
\caption{Dependencies of the survival model performance (the C-index) on the
proportion of uncensored data for the Beran estimator (the left graph) and the
Cox model (the right graph) when the MNIST-sin dataset is used}%
\label{f:mnistsin_c_c_ind}%
\end{center}
\end{figure}

\begin{figure}
[ptb]
\begin{center}
\includegraphics[
height=2.6792in,
width=5.3419in
]%
{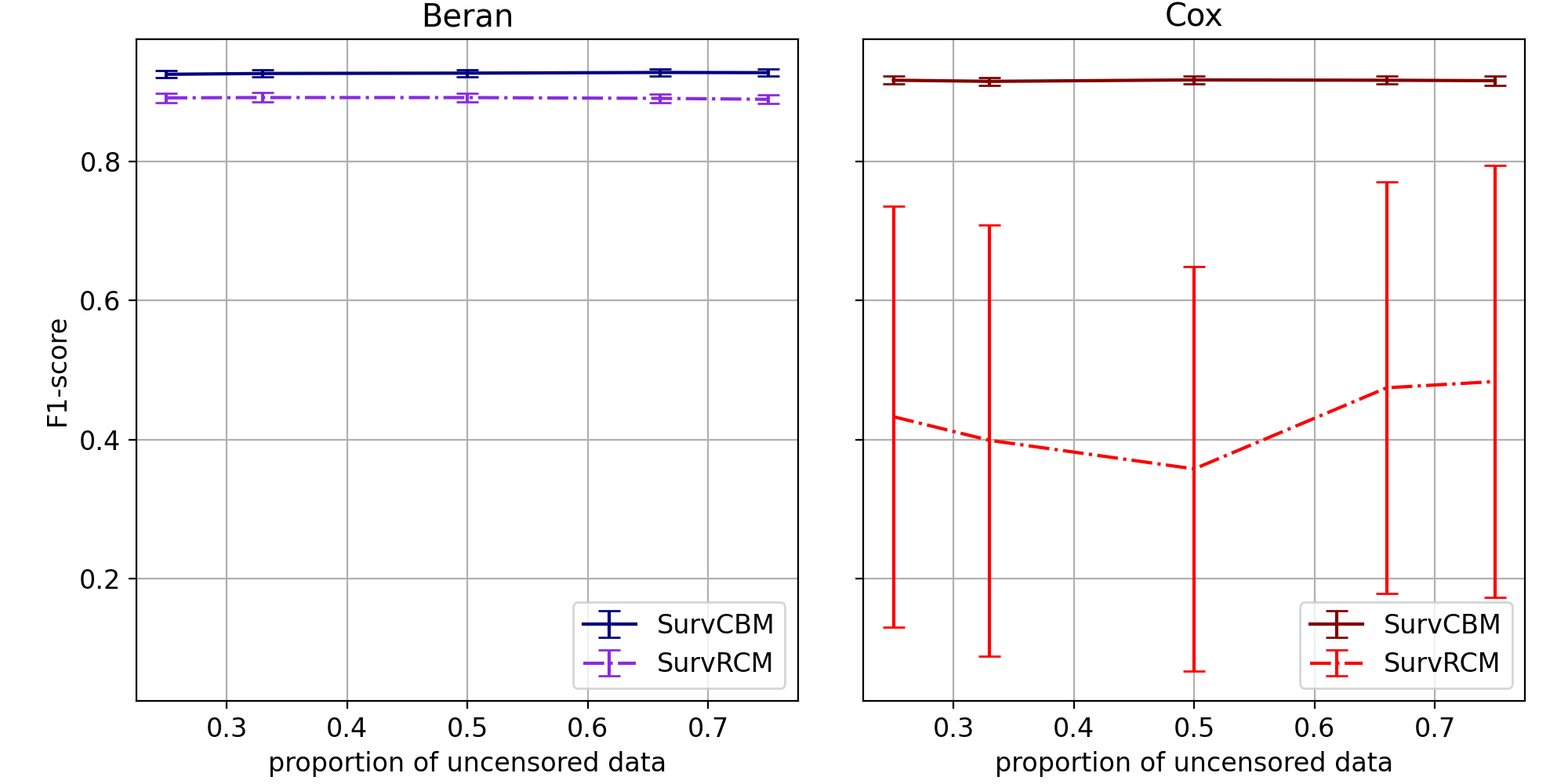}%
\caption{Dependencies of the survival model performance (the F1-measure) on
the proportion of uncensored data for the Beran estimator (the left graph) and
the Cox model (the right graph) when the MNIST-sin dataset is used}%
\label{f:mnistsin_c_f1}%
\end{center}
\end{figure}

Results depicted in Figs. \ref{f:mnistsin_c_c_ind} and \ref{f:mnistsin_c_f1}
are consistent with the similar results shown in Figs.
\ref{f:N_MNIST_SIN_C_IND} and \ref{f:mnistsin_n_f1}, respectively. We again
observe that the C-index increases with the number of training instances.
However, this increase is observed only for SurvCBM when the Beran estimator
is used. At the same time, there is almost no improvement in the concept
classification performance (the F1-measure) as the proportion of uncensored
observations increases (see Fig. \ref{f:mnistsin_c_f1}).%

%

\subsubsection{CIFAR-10}

\begin{figure}
[ptb]
\begin{center}
\includegraphics[
height=2.7311in,
width=5.4466in
]%
{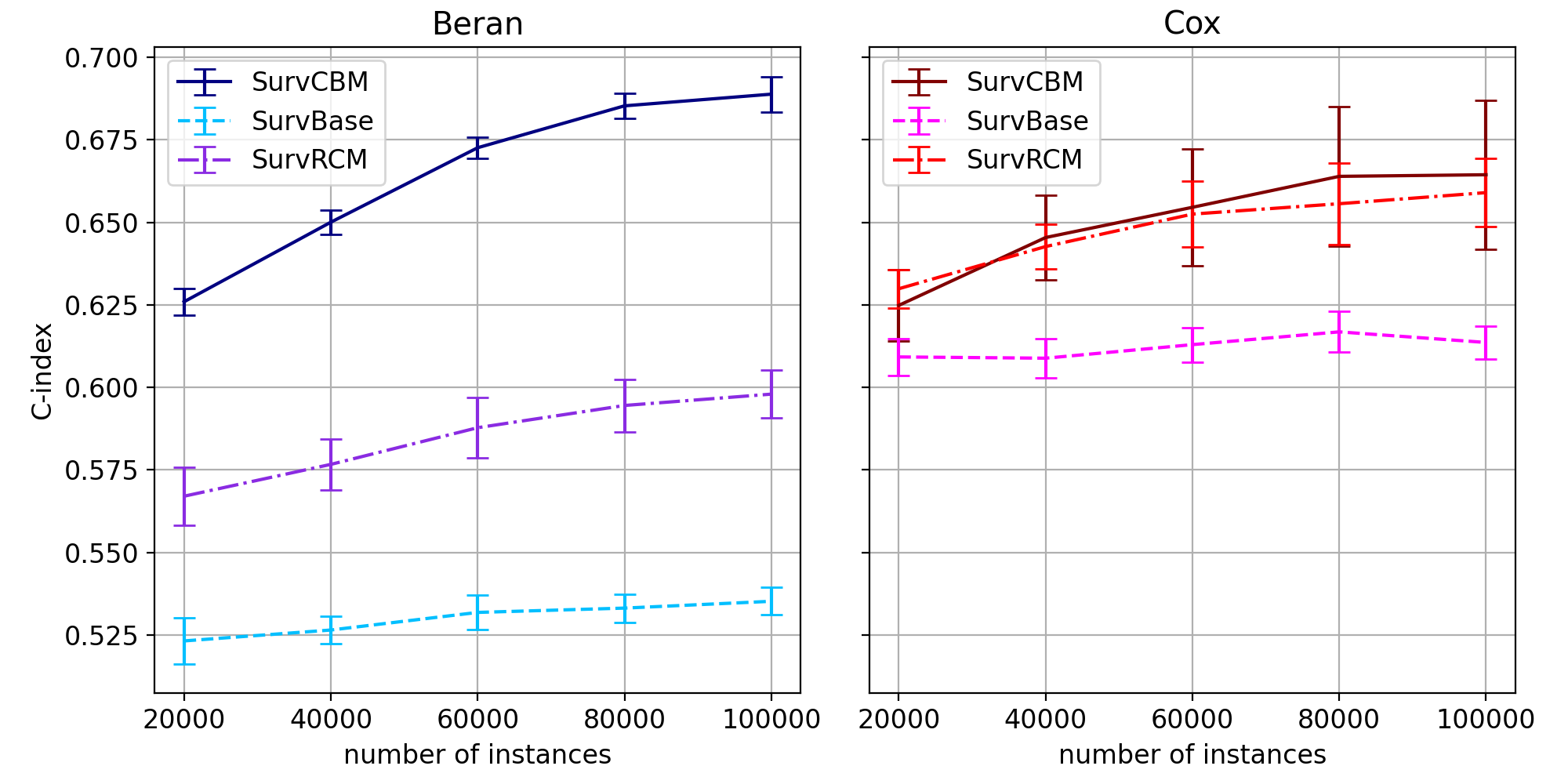}%
\caption{Dependencies of the survival model performance (the C-index) on the
number of training instances for the Beran estimator (the left graph) and the
Cox model (the right graph) when the CIFAR-10 dataset is used}%
\label{f:cifar_n_c_ind}%
\end{center}
\end{figure}

\begin{figure}
[ptb]
\begin{center}
\includegraphics[
height=2.7397in,
width=5.4639in
]%
{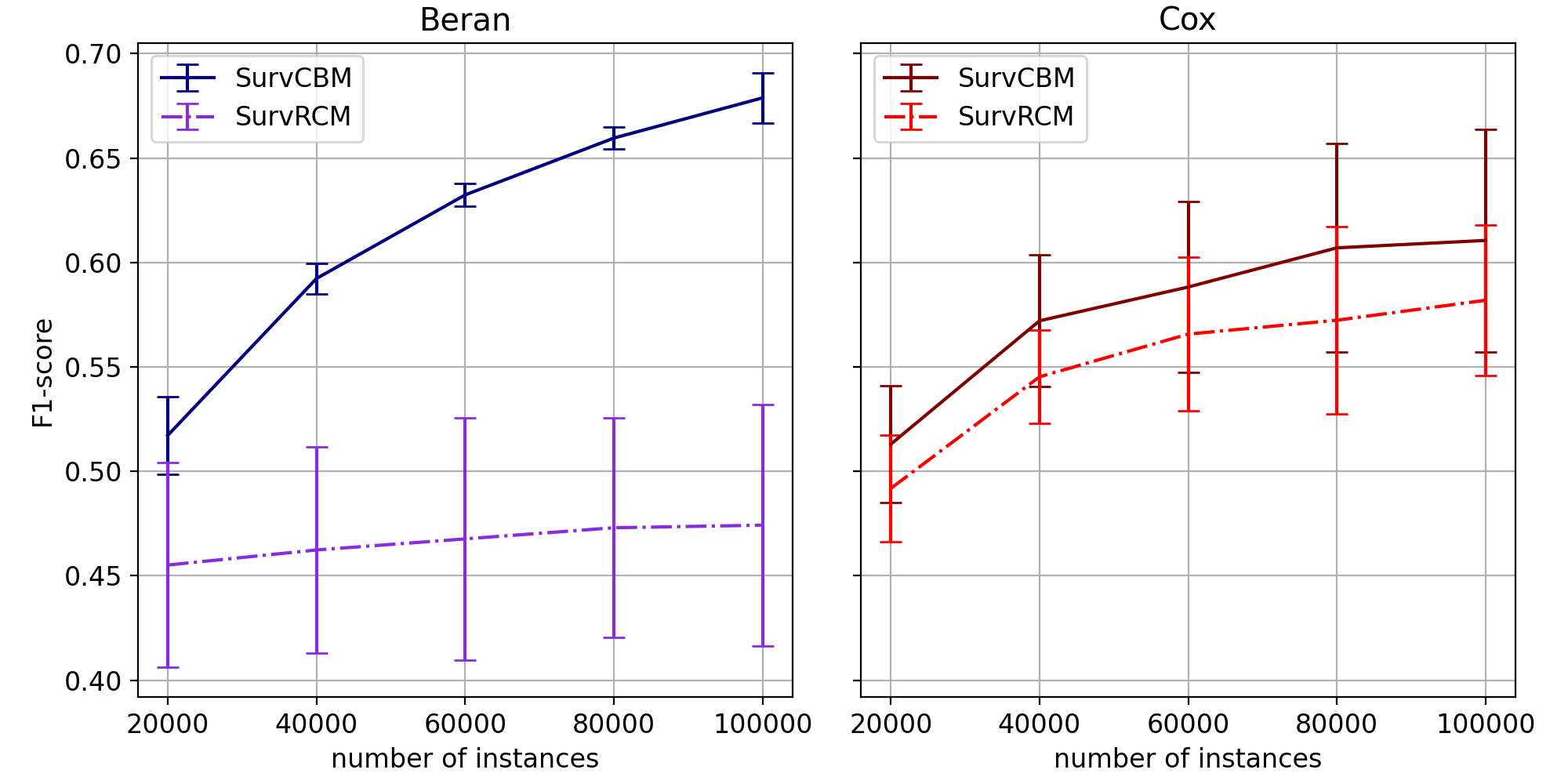}%
\caption{Dependencies of the survival model performance (the F1-measure) on
the number of training instances for the Beran estimator (the left graph) and
the Cox model (the right graph) when the CIFAR-10 dataset is used}%
\label{f:cifar_n_f1}%
\end{center}
\end{figure}

Similar numerical experiments are performed with the CIFAR-10 dataset. Figs.
\ref{f:cifar_n_c_ind} and \ref{f:cifar_n_f1} demonstrate that the more complex
data structure and principles of concept formation lead to worse results. The
complex structure of the dataset does not allow satisfactory results to be
obtained using the Cox model, even when the Weibull distribution is used to
generate the event times. At the same time, SurvCBM based on the Beran
estimator demonstrates superior results compared to the other models.%

%

\begin{figure}
[ptb]
\begin{center}
\includegraphics[
height=2.8089in,
width=5.6022in
]%
{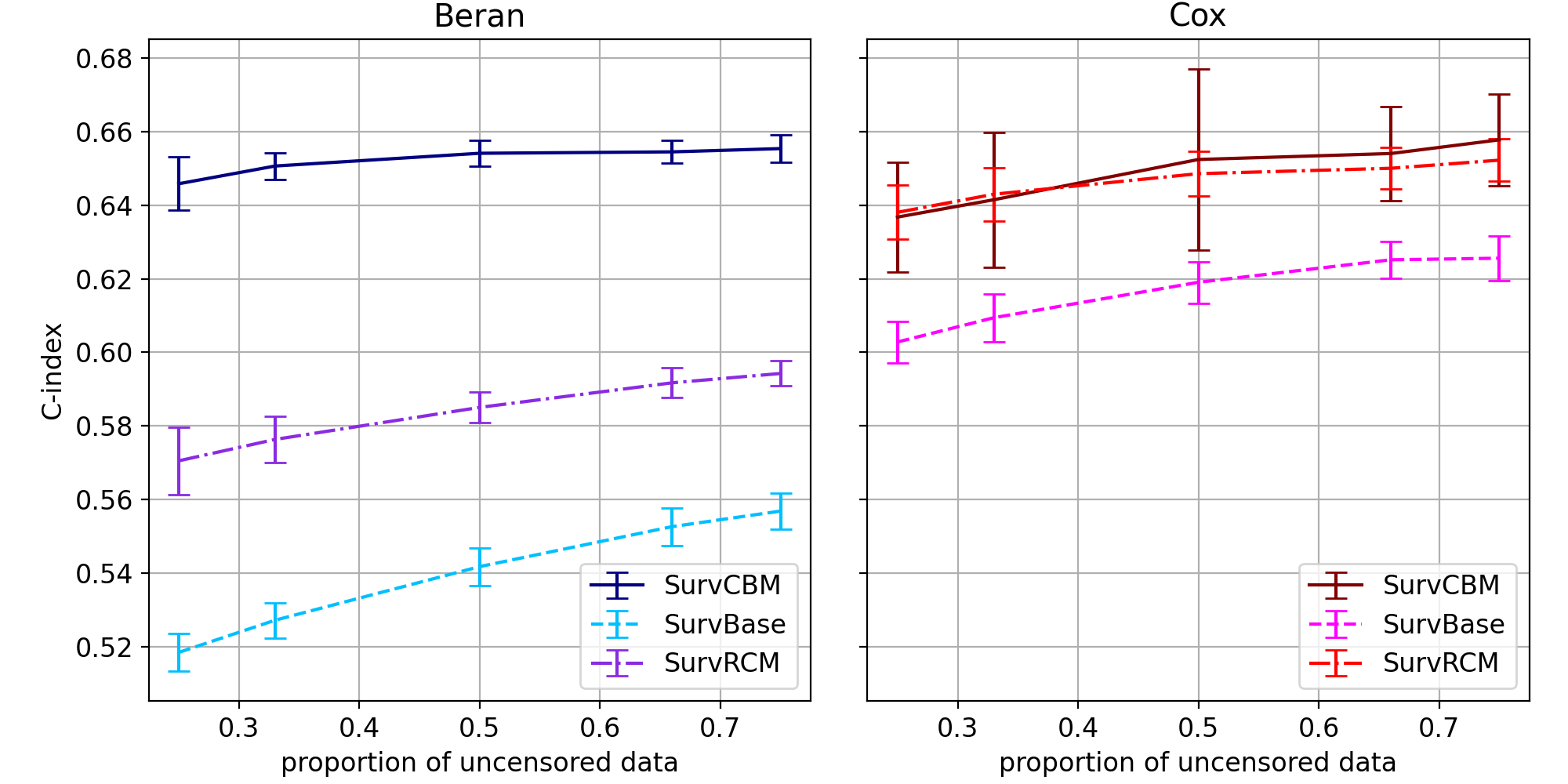}%
\caption{Dependencies of the survival model performance (the C-index) on the
proportion of uncensored data for the Beran estimator (the left graph) and the
Cox model (the right graph) when the CIFAR-10 dataset is used}%
\label{f:cifar_c_c_ind}%
\end{center}
\end{figure}

\begin{figure}
[ptb]
\begin{center}
\includegraphics[
height=2.8253in,
width=5.6368in
]%
{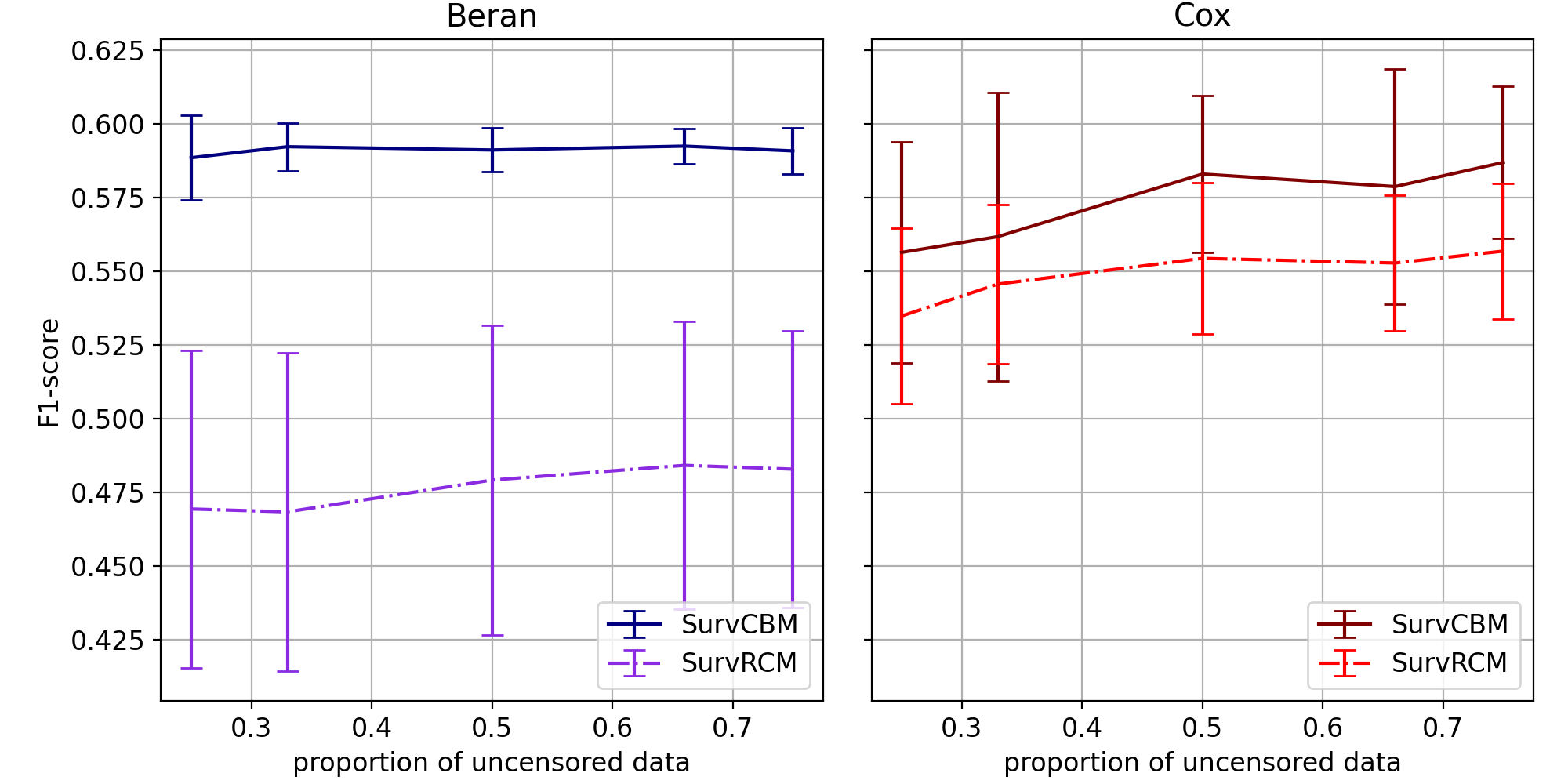}%
\caption{Dependencies of the survival model performance (the F1-measure) on
the proportion of uncensored data for the Beran estimator (the left graph) and
the Cox model (the right graph) when the CIFAR-10 dataset is used}%
\label{f:cifar_c_f1}%
\end{center}
\end{figure}

The same results are obtained when the F1-measures, as functions of the
uncensored data proportion, are studied. These results are shown in Figs.
\ref{f:cifar_c_c_ind} and \ref{f:cifar_c_f1}. At first glance, SurvCBM
achieves superior results when the Beran estimator is used. However, the
absolute values of the accuracy measures are quite low.%

%

\section{SurvCBM as an interpretation tool}

Depending on the survival model used in SurvCBM, we consider two methods for
interpreting predictions which are in the form of SFs.

The first method is applied when the Beran estimator is used in SurvCBM. The
goal of this method is to explain which concepts significantly impact the
predicted SF $S(t\mid \mathbf{x})$ for a new instance $\mathbf{x}$. A key
element of this interpretation is the predicted SF itself. However,
determining a direct relationship between concepts and the SF is challenging.
To address this, the following approach is proposed for interpreting the
predicted SF.

The model selects instances from the training set that are closest to the
instance being explained in the concept space. Closeness is defined as the
distance between the predicted SF of the explainable instance and the SFs of
other instances from the training data. The number of concepts that coincide
with those of the instance being explained, among the nearest instances,
determines the degree of importance of these concepts in the prediction
corresponding to the eplainable example. The more concepts that coincide, the
greater their importance. This interpretation can be viewed within the
framework of example-based explanations \cite{poche2023natural}.

For the MNIST dataset, the parameters $\mathbf{b}=(\mathbf{b}_{1}%
,\mathbf{b}_{2},\mathbf{b}_{3},\mathbf{b}_{4})\in \mathbb{R}^{40}$ in the
generation process (\ref{Weibull}) are set as follows: $\mathbf{b}%
_{1}=(0.5,...,0.5)\in \mathbb{R}^{10}$, $\mathbf{b}_{2}=(1.5,...,1.5)\in
\mathbb{R}^{10}$, $\mathbf{b}_{3}=(0.0001,...,0.0001)\in \mathbb{R}^{10}$,
$\mathbf{b}_{4}=(0.001,...,0.001)\in \mathbb{R}^{10}$. This indicates that the
first and second concepts have significant importance, while the third and
fourth concepts are relatively unimportant. Additionally, the parameter values
in (\ref{Weibull}) are set as $\lambda=10^{-4}$ and $\nu=2$. For the CIFAR-10
dataset, the parameters $\mathbf{b}=(\mathbf{b}_{1},\mathbf{b}_{2}%
,\mathbf{b}_{3},\mathbf{b}_{4})\in \mathbb{R}^{17}$ are defined as:
$\mathbf{b}_{1}=(0.001,...,0.001)\in \mathbb{R}^{5}$, $\mathbf{b}%
_{2}=(0.0001,...,0.0001)\in \mathbb{R}^{5}$, $\mathbf{b}_{3}=(3,...,3)\in
\mathbb{R}^{5}$, $\mathbf{b}_{4}=(5,5)\in \mathbb{R}^{2}$. Here, the third and
fourth concepts are important, while the first and second concepts are
unimportant. The parameters in (\ref{Weibull}) are set as $\lambda=10^{-3}$
and $\nu=4$.

To improve the classification performance and interpretability of the model,
we propose learning the parameter $\tau$ of the kernel in the Beran estimator
for each value of concepts. Weights $\alpha$ in (\ref{Weight2}) are of the
form:
\begin{equation}
\alpha(\mathbf{p},\mathbf{p}^{(k)})=\text{\textrm{softmax}}\left(  -\sum
_{i=1}^{m}\sum_{j=1}^{k_{i}}\frac{\left(  p_{i,j}-p_{i,j}^{(k)}\right)  ^{2}%
}{\tau_{i,j}}\right)  ,
\end{equation}
where $p_{i,j}$ is the $j$-th value logit of the $i$-th concept for the
explainable instance; $p_{i,j}^{(k)}$ is the $j$-th value logit of the $i$-th
concept for the $k$-th instance; $\tau_{i,j}$ is the trainable parameter.

This means that the number of training parameters $\tau_{i,j}$ is equal to $M$.

\begin{figure}
[ptb]
\begin{center}
\includegraphics[
height=3.1886in,
width=3.7533in
]%
{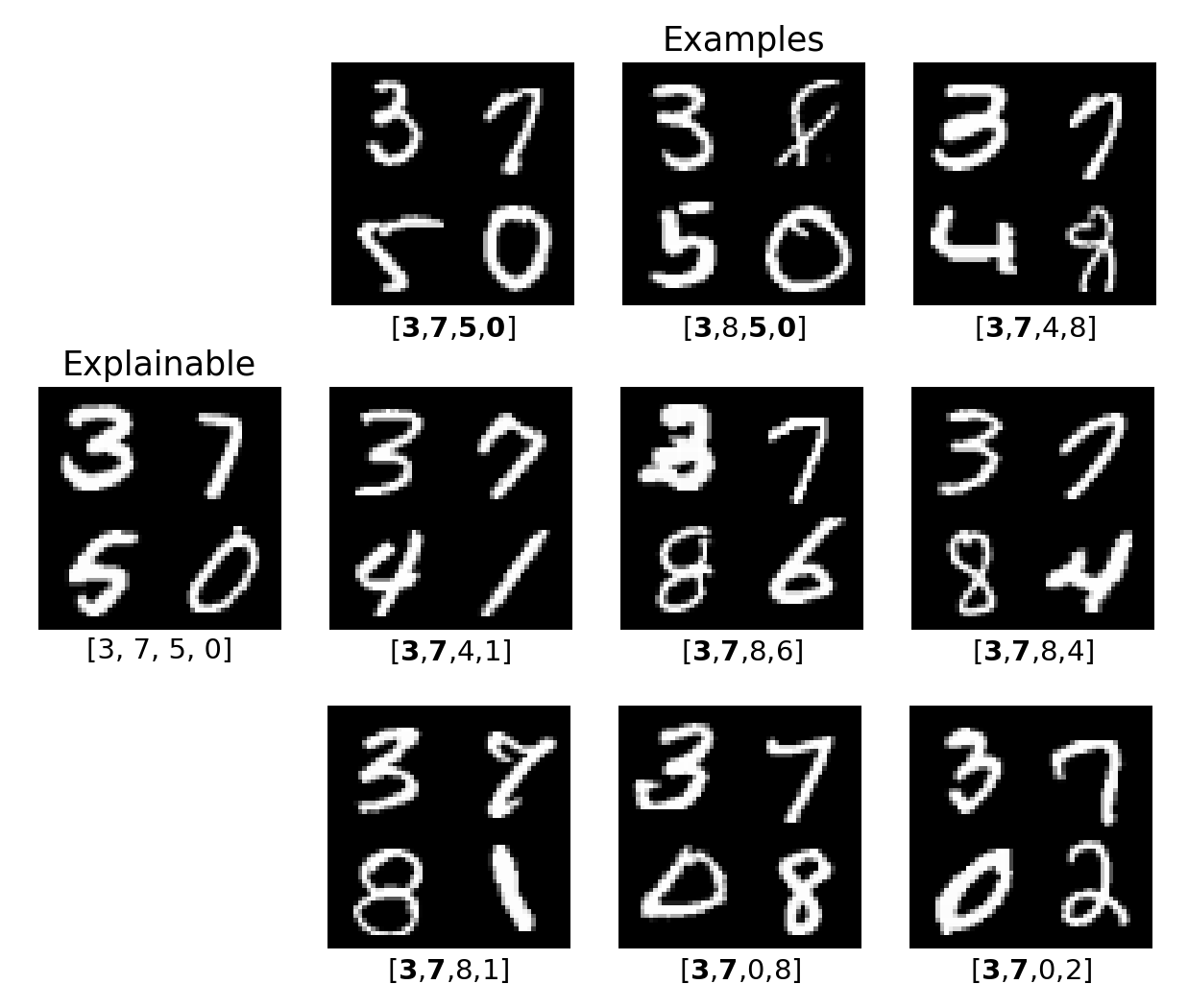}%
\caption{Examples of nearest neighbors for the explainable instance derived
from the MNIST dataset based on the Beran estimator}%
\label{f:explanation_mnist_beran_2}%
\end{center}
\end{figure}

\begin{figure}
[ptb]
\begin{center}
\includegraphics[
height=3.3338in,
width=3.8035in
]%
{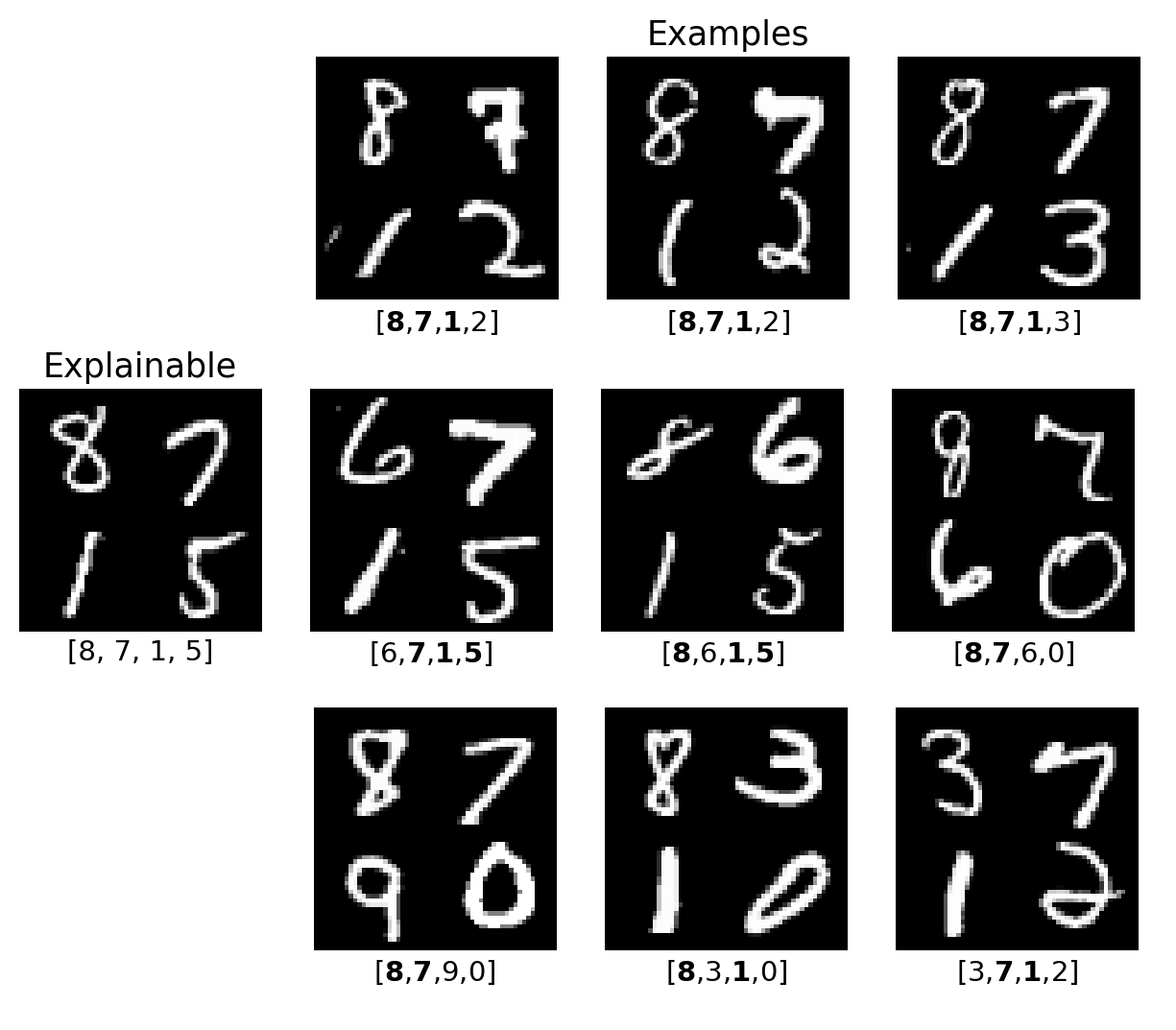}%
\caption{Examples of nearest neighbors for the explainable instance derived
from the MNIST dataset based on the Beran estimator}%
\label{f:explanation_mnist_beran_3}%
\end{center}
\end{figure}

Figs. \ref{f:explanation_mnist_beran_2} and \ref{f:explanation_mnist_beran_3}
illustrate examples of nearest instances constructed from the MNIST dataset.
In particular, it can be seen from Fig. \ref{f:explanation_mnist_beran_2} that
the value 3 of the first concept occurs in 9 out of 9 instances, the value 7
of the second concept occurs in 8 instances, the value 5 of the third concept
occurs in 2 instances, and the value 0 of the fourth concept occurs in 2
instances. This implies that the first concept is the most important for the
prediction of the explainable instance. The second concept is also important,
whereas the third and fourth concepts have only a slight impac t on the
prediction. The similar conclusion can be done by analyzing the example shown
in Fig. \ref{f:explanation_mnist_beran_3} where the value 8 of the first
concept and value 7 of the second concept occur in 7 nearest instances. This
implies that the first and the second concepts are the most important.%

%

\begin{figure}
[ptb]
\begin{center}
\includegraphics[
height=3.8899in,
width=4.427in
]%
{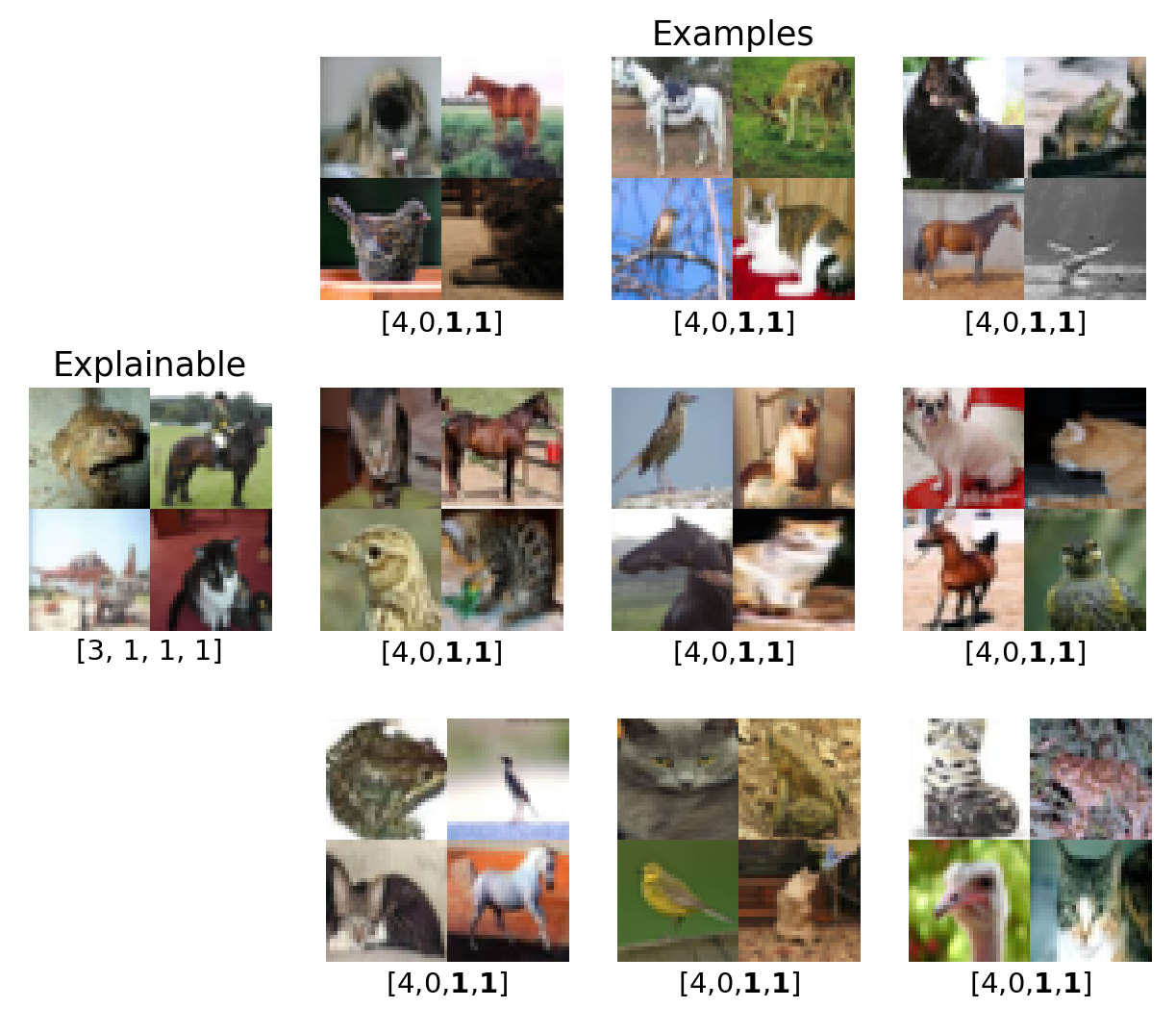}%
\caption{Examples of nearest neighbors for the explainable instance
constructed from the CIFAR-10 dataset based on the Beran estimator}%
\label{f:explanation_cifar_beran}%
\end{center}
\end{figure}

Fig. \ref{f:explanation_cifar_beran} illustrates an example of nearest
instances constructed from the CIFAR-10 datasets. It can be clearly seen from
Fig. \ref{f:explanation_cifar_beran} that the value 1 of the third concept and
the value 1 of the fourth concept are present in all nearest instances. This
indicates that the number of flying objects (the third concept) and the
presence of a cat in the image (the fourth concept) are important for
explaining the predictions of the explainable instance. Additionally, we
observe that all nearest instances share the same values for the first and
second concepts. However, these values do not match the corresponding concept
values of the explainable instance. Therefore, these concepts are not
considered important for the prediction.%

\begin{figure}
[ptb]
\begin{center}
\includegraphics[
height=1.0534in,
width=5.4648in
]%
{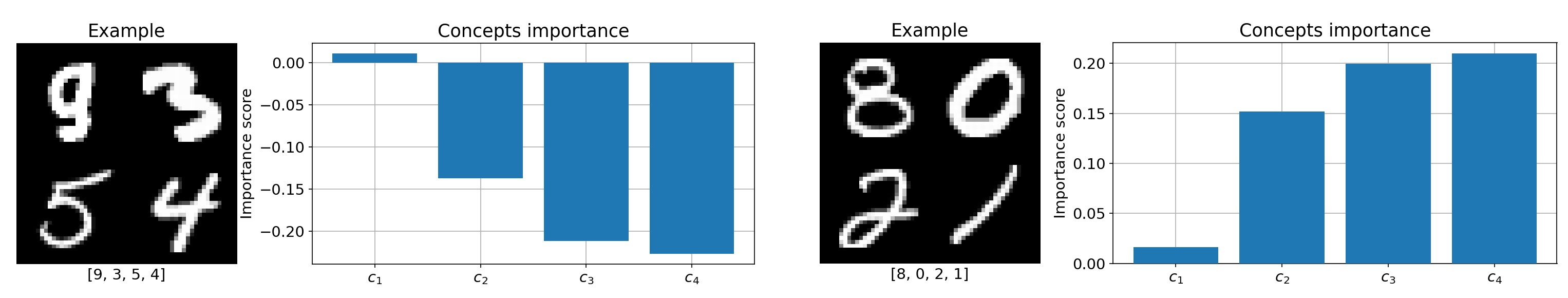}%
\caption{Two examples of the concept importance for explainable instances
constructed from the MNIST dataset based on the Cox model}%
\label{f:explanation_mnist_cox}%
\end{center}
\end{figure}

\begin{figure}
[ptb]
\begin{center}
\includegraphics[
height=1.1729in,
width=5.2638in
]%
{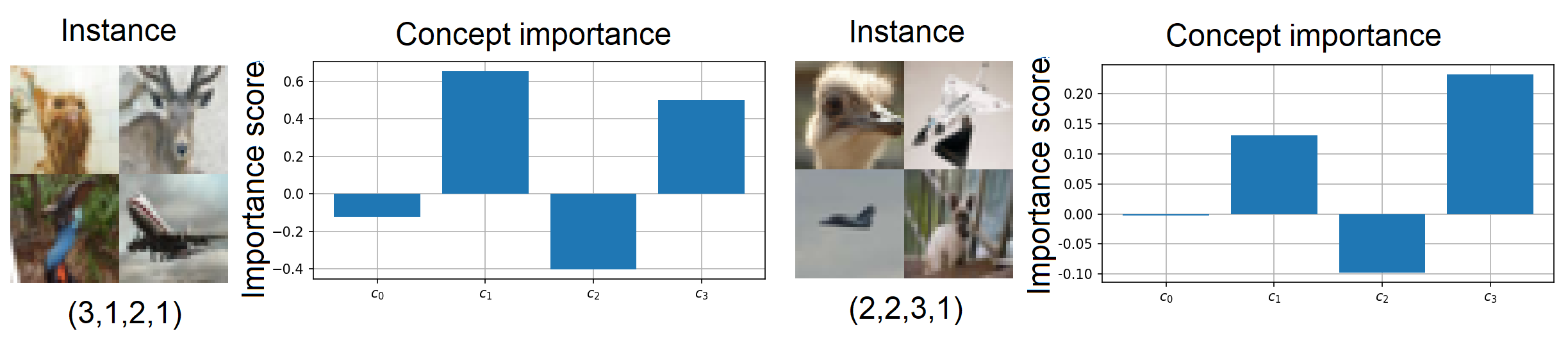}%
\caption{Two examples of the concept importance for explainable instances
constructed from the CIFAR-10 dataset based on the Cox model}%
\label{f:explanation_cifar_cox}%
\end{center}
\end{figure}

The second method is based on the linear relationship underlying the Cox
model. Instead of using logits to interpret predictions from the Cox model, we
transform the logits into probabilities%
\[
\mathbf{\pi}=(\mathbf{\pi}_{1},...,\mathbf{\pi}_{m})=(\pi_{i,1},...,\pi
_{i,k_{i}},i=1,...,m),
\]
using the softmax operator. Applying the trained regression coefficients
$\mathbf{b}$ from (\ref{Cox_SF}) and the probabilities $\mathbf{\pi}$ of the
concepts corresponding to the explainable instance, we can determine the
importance value of each concept as its contribution into the linear sum
$\mathbf{b}^{\mathrm{T}}\mathbf{\pi}$. In other words, we compute the product
$\mathbf{b}_{i}^{\mathrm{T}}\mathbf{\pi}_{i}$ to quantify the $i$-th concept
contribution, where $\mathbf{b}_{i}$ is the $i$-th part of the vector
$\mathbf{b}$ corresponding to the vector $\mathbf{\pi}_{i}=(\pi_{i,1}%
,...,\pi_{i,k_{i}})$. Figs. \ref{f:explanation_mnist_cox} and
\ref{f:explanation_cifar_cox} illustrate the explainable instances constructed
from the MNIST dataset and from the CIFAR-10 dataset, respectively, along with
their concept importance values. It should be noted that the event times are
generated in accordance with vectors of coefficients $(0,7,10,14)$ for the
MNIST dataset and $(-0.7,1.5,-2.0,5.0)$ for the CIFAR-10 dataset.%

%

\section{Conclusion}

Two concept-based survival learning models have been proposed and compared.
Based on numerous numerical experiments and the architectures of the models,
SurvCBM outperforms SurvRCM in all cases. Therefore, we focus primarily on the
key features of SurvCBM. It is worth noting that the proposed model represents
the first attempt to integrate concept-based information into survival
analysis. Through extensive numerical experiments, this approach has
demonstrated that concept information plays a crucial role in improving both
classification and regression accuracy.

SurvCBM is based on a simple implementation of the Beran estimator, which uses
only one training parameter for the Gaussian kernel. A potential extension
could involve incorporating additional training parameters into the Beran
estimator to enhance its generalization capabilities. Another promising
direction is to replace the Beran estimator and the Cox model with neural
networks designed to perform the same survival analysis tasks. These
modifications represent important avenues for future research.

Another distinctive feature of SurvCBM is its representation of concepts in
categorical form. However, it would be valuable to explore concepts that
represent time-to-event or other continuous values. This representation is
particularly relevant when analyzing the reliability of complex systems
composed of multiple components with different reliability characteristics.
Such a modification would enable the interpretation of system predictions in
terms of unreliable components and could serve as another direction for
further research.

The proposed models assume that each instance is described by a set of
predefined concepts available a priori. However, in many real-world scenarios,
concept information may not be readily available. Therefore, developing
approaches to automatically discover concepts from data is an important
challenge. While this problem has been addressed in \cite{rao2024discover}, it
has not yet been explored within the framework of survival analysis, making it
a promising direction for future work.

An important task in many applications, particularly in medicine, is to
incorporate expert rules in the form of logical functions of concepts into the
learning and inference processes. This approach can be regarded as a
combination of inductive and deductive learning in CBL. It has been studied in
\cite{Konstantinov-Utkin-24c,Konstantinov-Utkin-24}. However, it can be
extended to the case of survival analysis, which represents another promising
direction for research.

\bibliographystyle{unsrt}
\bibliography{Autoencoder,Classif_bib,Cluster_bib,Concept,Explain,Explain_med,MIL,MYBIB,Survival_analysis}

\begin{thebibliography}{10}

\bibitem{Gupta-Narayanan-24}
A.~Gupta and P.J. Narayanan.
\newblock A survey on concept-based approaches for model improvement.
\newblock arXiv:2403.14566, Mar 2024.

\bibitem{kim2018interpretability}
Been Kim, M.~Wattenberg, J.~Gilmer, C.~Cai, J.~Wexler, F.~Viegas, et~al.
\newblock Interpretability beyond feature attribution: Quantitative testing
  with concept activation vectors (tcav).
\newblock In {\em International conference on machine learning}, pages
  2668--2677. PMLR, 2018.

\bibitem{lage2020learning}
I.~Lage and F.~Doshi-Velez.
\newblock Learning interpretable concept-based models with human feedback.
\newblock arXiv:2012.02898, Dec 2020.

\bibitem{wang2023learning}
Bowen Wang, Liangzhi Li, Y.~Nakashima, and H.~Nagahara.
\newblock Learning bottleneck concepts in image classification.
\newblock In {\em Proceedings of the IEEE/CVF Conference on Computer Vision and
  Pattern Recognition}, pages 10962--10971, 2023.

\bibitem{xu2023statistically}
Kaiwen Xu, Kazuto Fukuchi, Youhei Akimoto, and Jun Sakuma.
\newblock Statistically significant concept-based explanation of image
  classifiers via model knockoffs.
\newblock In {\em Proceedings of the Thirty-Second International Joint
  Conference on Artificial Intelligence}, pages 519--526, 2023.

\bibitem{yeh2020completeness}
Chih-Kuan Yeh, Been Kim, S.~Arik, Chun-Liang Li, T.~Pfister, and P.~Ravikumar.
\newblock On completeness-aware concept-based explanations in deep neural
  networks.
\newblock In {\em Advances in neural information processing systems},
  volume~33, pages 20554--20565, 2020.

\bibitem{Guidotti-2019}
R.~Guidotti, A.~Monreale, S.~Ruggieri, F.~Turini, F.~Giannotti, and
  D.~Pedreschi.
\newblock A survey of methods for explaining black box models.
\newblock {\em ACM computing surveys}, 51(5):93, 2019.

\bibitem{Liang-etal-2021}
Y.~Liang, S.~Li, C.~Yan, M.~Li, and C.~Jiang.
\newblock Explaining the black-box model: A survey of local interpretation
  methods for deep neural networks.
\newblock {\em Neurocomputing}, 419:168--182, 2021.

\bibitem{Zhang-Tino-etal-2020}
Yu~Zhang, Peter Ti{\v{n}}o, Ale{\v{s}} Leonardis, and Ke~Tang.
\newblock A survey on neural network interpretability.
\newblock {\em IEEE Transactions on Emerging Topics in Computational
  Intelligence}, 5(5):726--742, 2021.

\bibitem{poeta2023concept}
E.~Poeta, G.~Ciravegna, E.~Pastor, T.~Cerquitelli, and E.~Baralis.
\newblock Concept-based explainable artificial intelligence: A survey.
\newblock arXiv:2312.12936, May 2023.

\bibitem{hu2024editable}
Lijie Hu, Chenyang Ren, Zhengyu Hu, Cheng-Long Wang, and Di~Wang.
\newblock Editable concept bottleneck models.
\newblock arXiv:2405.15476, May 2024.

\bibitem{koh2020concept}
Pang~Wei Koh, Thao Nguyen, Yew~Siang Tang, S.~Mussmann, E.~Pierson, Been Kim,
  and Percy Liang.
\newblock Concept bottleneck models.
\newblock In {\em International conference on machine learning}, pages
  5338--5348. PMLR, 2020.

\bibitem{chauhan2023interactive}
K.~Chauhan, R.~Tiwari, J.~Freyberg, P.~Shenoy, and K.~Dvijotham.
\newblock Interactive concept bottleneck models.
\newblock In {\em Proceedings of the AAAI Conference on Artificial
  Intelligence}, volume~37, pages 5948--5955, 2023.

\bibitem{Sarkar_2022_CVPR}
A.~Sarkar, D.~Vijaykeerthy, A.~Sarkar, and V.N. Balasubramanian.
\newblock A framework for learning ante-hoc explainable models via concepts.
\newblock In {\em Proceedings of the IEEE/CVF Conference on Computer Vision and
  Pattern Recognition (CVPR)}, pages 10286--10295, June 2022.

\bibitem{kim2023probabilistic}
Eunji Kim, Dahuin Jung, Sangha Park, Siwon Kim, and Sungroh Yoon.
\newblock Probabilistic concept bottleneck models.
\newblock In {\em International Conference on Machine Learning}, pages
  16521--16540. PMLR, 2023.

\bibitem{ismail2023concept}
A.A. Ismail, J.~Adebayo, H.C. Bravo, S.~Ra, and Kyunghyun Cho.
\newblock Concept bottleneck generative models.
\newblock In {\em Proceedings of ICML 2023. Workshop on Deployment Challenges
  for Generative AI}, pages 1--10, 2023.

\bibitem{kazmierczak2024clipqda}
R.~Kazmierczak, E.~Berthier, G.~Frehse, and G.~Franchi.
\newblock {CLIP}-{QDA}: An explainable concept bottleneck model.
\newblock {\em Transactions on Machine Learning Research}, 2024.

\bibitem{marconato2022glancenets}
E.~Marconato, A.~Passerini, and S.~Teso.
\newblock Glancenets: Interpretable, leak-proof concept-based models.
\newblock In {\em Advances in Neural Information Processing Systems},
  volume~35, pages 21212--21227, 2022.

\bibitem{vandenhirtz2024stochastic}
M.~Vandenhirtz, S.~Laguna, R.~Marcinkevi{\v{c}}s, and J.E. Vogt.
\newblock Stochastic concept bottleneck models.
\newblock In {\em ICML 2024 Workshop on Structured Probabilistic Inference {\&}
  Generative Modeling}, 2024.

\bibitem{zarlenga2023tabcbm}
M.E. Zarlenga, Z.~Shams, M.E. Nelson, B.~Kim, and M.~Jamnik.
\newblock Tab{CBM}: Concept-based interpretable neural networks for tabular
  data.
\newblock {\em Transactions on Machine Learning Research}, 2023.

\bibitem{Hosmer-Lemeshow-May-2008}
D.~Hosmer, S.~Lemeshow, and S.~May.
\newblock {\em Applied Survival Analysis: Regression Modeling of Time to Event
  Data}.
\newblock John Wiley \& Sons, New Jersey, 2008.

\bibitem{forest2024interpretable}
F.~Forest, K.~Rombach, and O.~Fink.
\newblock Interpretable prognostics with concept bottleneck models.
\newblock arXiv:2405.17575, May 2024.

\bibitem{Cox-1972}
D.R. Cox.
\newblock Regression models and life-tables.
\newblock {\em Journal of the Royal Statistical Society, Series B
  (Methodological)}, 34(2):187--220, 1972.

\bibitem{Kovalev-Utkin-Kasimov-20a}
M.S. Kovalev, L.V. Utkin, and E.M. Kasimov.
\newblock Surv{LIME}: A method for explaining machine learning survival models.
\newblock {\em Knowledge-Based Systems}, 203:106164, 2020.

\bibitem{Beran-81}
R.~Beran.
\newblock Nonparametric regression with randomly censored survival data.
\newblock Technical report, University of California, Berkeley, 1981.

\bibitem{Utkin-Eremenko-Konstantinov-24}
L.V. Utkin, D.Y. Eremenko, and A.V. Konstantinov.
\newblock Surv{B}e{X}: an explanation method of the machine learning survival
  models based on the beran estimator.
\newblock {\em International Journal of Data Science and Analytics}, pages
  1--26, 2024.

\bibitem{Sheth-Kahou-23}
Ivaxi Sheth and Samira Ebrahimi~Kahou.
\newblock Auxiliary losses for learning generalizable concept-based models.
\newblock {\em Advances in Neural Information Processing Systems}, 36, 2024.

\bibitem{hu2024semi}
Lijie Hu, Tianhao Huang, Huanyi Xie, Chenyang Ren, Zhengyu Hu, Lu~Yu, and
  Di~Wang.
\newblock Semi-supervised concept bottleneck models.
\newblock arXiv:2406.18992, Jun 2024.

\bibitem{oikarinen2023label}
T.~Oikarinen, S.~Das, L.M. Nguyen, and Tsui-Wei Weng.
\newblock Label-free concept bottleneck models.
\newblock arXiv:2304.06129, Apr 2023.

\bibitem{Wang-Junlin-Chen-24}
Hongmei Wang, Junlin Hou, and Hao Chen.
\newblock Concept complement bottleneck model for interpretable medical image
  diagnosis.
\newblock arXiv:2410.15446, Oct 2024.

\bibitem{schrodi2024concept}
S.~Schrodi, J.~Schur, M.~Argus, and T.~Brox.
\newblock Concept bottleneck models without predefined concepts.
\newblock arXiv:2407.03921, Jul 2024.

\bibitem{yuksekgonul2022post}
Mert Yuksekgonul, Maggie Wang, and James Zou.
\newblock Post-hoc concept bottleneck models.
\newblock In {\em ICLR 2022 Workshop on PAIR
  $\{$$\backslash$textasciicircum$\}$ 2Struct: Privacy, Accountability,
  Interpretability, Robustness, Reasoning on Structured Data}, 2022.

\bibitem{shang2024incremental}
Chenming Shang, Shiji Zhou, Hengyuan Zhang, Xinzhe Ni, Yujiu Yang, and Yuwang
  Wang.
\newblock Incremental residual concept bottleneck models.
\newblock In {\em Proceedings of the IEEE/CVF Conference on Computer Vision and
  Pattern Recognition}, pages 11030--11040, 2024.

\bibitem{dominici2024anycbms}
Gabriele Dominici, Pietro Barbiero, Francesco Giannini, Martin Gjoreski, and
  Marc Langhenirich.
\newblock Any{CBM}s: How to turn any black box into a concept bottleneck model.
\newblock arXiv:2405.16508, May 2024.

\bibitem{Aysel-etal-25}
H.I. Aysel, Xiaohao Cai, and A.~Prugel-Bennett.
\newblock Concept-based explainable artificial intelligence: Metrics and
  benchmarks.
\newblock arXiv:2501.19271, Jan 2025.

\bibitem{lee2023neural}
Jae~Hee Lee, S.~Lanza, and S.~Wermter.
\newblock From neural activations to concepts: A survey on explaining concepts
  in neural networks.
\newblock arXiv:2310.11884, Oct 2023.

\bibitem{mahinpei2021promises}
A.~Mahinpei, J.~Clark, I.~Lage, F.~Doshi-Velez, and Weiwei Pan.
\newblock Promises and pitfalls of black-box concept learning models.
\newblock arXiv:2106.13314, Jun 2021.

\bibitem{Wang-Li-Reddy-2019}
P.~Wang, Y.~Li, and C.K. Reddy.
\newblock Machine learning for survival analysis: A survey.
\newblock {\em {ACM} Computing Surveys (CSUR)}, 51(6):1--36, 2019.

\bibitem{salerno2023high}
Stephen Salerno and Yi~Li.
\newblock High-dimensional survival analysis: Methods and applications.
\newblock {\em Annual review of statistics and its application}, 10:25--49,
  2023.

\bibitem{Wiegrebe:2024aa}
Simon Wiegrebe, Philipp Kopper, Raphael Sonabend, Bernd Bischl, and Andreas
  Bender.
\newblock Deep learning for survival analysis: a review.
\newblock {\em Artificial Intelligence Review}, 57(65):1--34, 2024.

\bibitem{chen2024introduction}
G.H. Chen.
\newblock An introduction to deep survival analysis models for predicting
  time-to-event outcomes.
\newblock {\em Foundations and Trends{\textregistered} in Machine Learning},
  17(6):921--1100, 2024.

\bibitem{EmmertStreib-Dehmer-19}
F.~Emmert-Streib and M.~Dehmer.
\newblock Introduction to survival analysis in practice.
\newblock {\em Machine Learning \& Knowledge Extraction}, 1:1013--1038, 2019.

\bibitem{Katzman-etal-2018}
J.L. Katzman, U.~Shaham, A.~Cloninger, J.~Bates, T.~Jiang, and Y.~Kluger.
\newblock Deepsurv: Personalized treatment recommender system using a {C}ox
  proportional hazards deep neural network.
\newblock {\em BMC medical research methodology}, 18(24):1--12, 2018.

\bibitem{Luck-etal-2017}
M.~Luck, T.~Sylvain, H.~Cardinal, A.~Lodi, and Y.~Bengio.
\newblock Deep learning for patient-specific kidney graft survival analysis.
\newblock arXiv:1705.10245, May 2017.

\bibitem{Nezhad-etal-2018}
M.Z. Nezhad, N.~Sadati, K.~Yang, and D.~Zhu.
\newblock A deep active survival analysis approach for precision treatment
  recommendations: Application of prostate cancer.
\newblock arXiv:1804.03280v1, April 2018.

\bibitem{ren2019deep}
Kan Ren, Jiarui Qin, Lei Zheng, Zhengyu Yang, Weinan Zhang, Lin Qiu, and Yong
  Yu.
\newblock Deep recurrent survival analysis.
\newblock In {\em Proceedings of the AAAI conference on artificial
  intelligence}, volume~33, pages 4798--4805, 2019.

\bibitem{Steingrimsson-Morrison-20}
J.A. Steingrimsson and S.~Morrison.
\newblock Deep learning for survival outcomes.
\newblock {\em Statistics in Medicine}, 39(17):2339--2349, 2020.

\bibitem{Tarkhan-etal-21}
A.~Tarkhan, N.~Simon, T.~Bengtsson, K.~Nguyen, and J.~Dai.
\newblock Survival prediction using deep learning.
\newblock In {\em Proceedings of AAAI Spring Symposium on Survival
  Prediction-Algorithms, Challenges and Applications}, volume 146, pages
  207--214. PMLR, 2021.

\bibitem{Yao-Zhu-Zhu-Huang-2017}
J.~Yao, X.~Zhu, F.~Zhu, and J.~Huang.
\newblock Deep correlational learning for survival prediction from
  multi-modality data.
\newblock In {\em Medical Image Computing and Computer--Assisted Intervention
  -- MICCAI 2017}, volume 10434 of {\em Lecture Notes in Computer Science},
  pages 406--414. Springer, Cham, 2017.

\bibitem{Zhong-Mueller-Wang-21}
Qixian Zhong, J.W. Mueller, and Jane-Ling Wang.
\newblock Deep extended hazard models for survival analysis.
\newblock In {\em Advances in Neural Information Processing Systems},
  volume~34, pages 15111--15124. Curran Associates, Inc., 2021.

\bibitem{Chatha-etal-22}
Prayag Chatha, Yixin Wang, Zhenke Wu, and J.~Regier.
\newblock Dynamic survival transformers for causal inference with electronic
  health records.
\newblock arXiv:2210.15417, Oct. 2022.

\bibitem{hu2021transformer}
Shi Hu, Egill Fridgeirsson, Guido van Wingen, and Max Welling.
\newblock Transformer-based deep survival analysis.
\newblock In {\em Survival Prediction-Algorithms, Challenges and Applications},
  pages 132--148. PMLR, 2021.

\bibitem{Li-Zhu-Yao-Huang-22}
Chunyuan Li, Xinliang Zhu, Jiawen Yao, and Junzhou Huang.
\newblock Hierarchical transformer for survival prediction using multimodality
  whole slide images and genomics.
\newblock In {\em The 26th International Conference on Pattern Recognition
  (ICPR)}, pages 4256--4262. IEEE Computer Society, 2022.

\bibitem{Lv-Lin-etal-22}
Zhilong Lv, Yuexiao Lin, Rui Yan, Ying Wang, and Fa~Zhang.
\newblock Transsurv: Transformer-based survival analysis model integrating
  histopathological images and genomic data for colorectal cancer.
\newblock {\em IEEE/ACM Transactions on Computational Biology and
  Bioinformatics}, pages 1--10, 2022.

\bibitem{Shen-liu-etal-22}
Yifan Shen, Li~liu, Zhihao Tang, Zongyi Chen, Guixiang Ma, Jiyan Dong,
  Xi~Zhang, Lin Yang, and Qingfeng Zheng.
\newblock Explainable survival analysis with convolution-involved vision
  transformer.
\newblock In {\em Proceedings of the AAAI Conference on Artificial Intelligence
  (AAAI-22)}, volume~36, pages 2207--2215, 2022.

\bibitem{tang2023explainable}
Zhihao Tang, Li~Liu, Zongyi Chen, Guixiang Ma, Jiyan Dong, Xujie Sun, Xi~Zhang,
  Chaozhuo Li, Qingfeng Zheng, Lin Yang, et~al.
\newblock Explainable survival analysis with uncertainty using
  convolution-involved vision transformer.
\newblock {\em Computerized Medical Imaging and Graphics}, 110:102302, 2023.

\bibitem{Wang-Sun-22}
Zifeng Wang and Jimeng Sun.
\newblock Survtrace: Transformers for survival analysis with competing events.
\newblock In {\em Proceedings of the 13th ACM International Conference on
  Bioinformatics, Computational Biology and Health Informatics}, pages 1--9,
  2022.

\bibitem{Li-Krivtsov-Arora-22}
Xingyu Li, V.~Krivtsov, and K.~Arora.
\newblock Attention-based deep survival model for time series data.
\newblock {\em Reliability Engineering and System Safety}, 217(108033):1--12,
  2022.

\bibitem{Sun-Dong-etal-21}
Zhaohong Sun, Wei Dong, Jinlong Shi, Kunlun He, and Zhengxing Huang.
\newblock Attention-based deep recurrent model for survival prediction.
\newblock {\em ACM Transactions on Computing for Healthcare}, 2(4):1--18, 2021.

\bibitem{Ibrahim-etal-2008}
N.A. Ibrahim, A.~Kudus, I.~Daud, and M.R.~Abu Bakar.
\newblock Decision tree for competing risks survival probability in breast
  cancer study.
\newblock {\em International Journal Of Biological and Medical Research},
  3(1):25--29, 2008.

\bibitem{Wright-etal-2017}
M.N. Wright, T.~Dankowski, and A.~Ziegler.
\newblock Unbiased split variable selection for random survival forests using
  maximally selected rank statistics.
\newblock {\em Statistics in Medicine}, 36(8):1272--1284, 2017.

\bibitem{Haarburger-etal-2018}
C.~Haarburger, P.~Weitz, O.~Rippel, and D.~Merhof.
\newblock Image-based survival analysis for lung cancer patients using {CNN}s.
\newblock arXiv:1808.09679v1, Aug 2018.

\bibitem{Widodo-Yang-2011}
A.~Widodo and B.-S. Yang.
\newblock Machine health prognostics using survival probability and support
  vector machine.
\newblock {\em Expert Systems with Applications}, 38(7):8430--8437, 2011.

\bibitem{Witten-Tibshirani-2010}
D.M. Witten and R.~Tibshirani.
\newblock Survival analysis with high-dimensional covariates.
\newblock {\em Statistical Methods in Medical Research}, 19(1):29--51, 2010.

\bibitem{Harrell-etal-1982}
F.~Harrell, R.~Califf, D.~Pryor, K.~Lee, and R.~Rosati.
\newblock Evaluating the yield of medical tests.
\newblock {\em Journal of the American Medical Association}, 247:2543--2546,
  1982.

\bibitem{Uno-etal-11}
H.~Uno, Tianxi Cai, M.J. Pencina, R.B. D'Agostino, and Lee-Jen Wei.
\newblock On the c-statistics for evaluating overall adequacy of risk
  prediction procedures with censored survival data.
\newblock {\em Statistics in medicine}, 30(10):1105--1117, 2011.

\bibitem{LeCun-etal-98}
Y.~LeCun, L.~Bottou, Y.~Bengio, and P.~Haffner.
\newblock Gradient-based learning applied to document recognition.
\newblock {\em Proceedings of the IEEE}, 86(11):2278--2324, 1998.

\bibitem{Krizhevsky-Hinton-2009}
A.~Krizhevsky and G.~Hinton.
\newblock Learning multiple layers of features from tiny images.
\newblock Technical Report~1, Computer Science Department, University of
  Toronto, 2009.

\bibitem{poche2023natural}
A.~Poch{\'e}, L.~Hervier, and M.-C. Bakkay.
\newblock Natural example-based explainability: a survey.
\newblock In {\em World Conference on eXplainable Artificial Intelligence},
  pages 24--47. Springer, 2023.

\bibitem{rao2024discover}
S.S. Rao, S.~Mahajan, M.~B{\"o}hle, and B.~Schiele.
\newblock Discover-then-name: Task-agnostic concept bottlenecks via automated
  concept discovery.
\newblock In {\em 18th European Conference on Computer Vision}, pages 444--461.
  Springer, 2024.

\bibitem{Konstantinov-Utkin-24c}
A.V. Konstantinov and L.V. Utkin.
\newblock An explicit concept-based approach for incorporating expert rules
  into machine learning models.
\newblock In {\em Proceedings of the Eighth International Scientific Conference
  Intelligent Information Technologies for Industry (IITI'24)}, volume~1 of
  {\em Lecture Notes in Networks and Systems, vol. 1209}, pages 153--162.
  Springer, Cham, 2024.

\bibitem{Konstantinov-Utkin-24}
A.V. Konstantinov and L.V. Utkin.
\newblock Incorporating expert rules into neural networks in the framework of
  concept-based learning.
\newblock arXiv:2402.14726, Feb 2024.

\end{thebibliography}

\end{document}